\documentclass[runningheads]{llncs}
\usepackage{bbding}
\usepackage[T1]{fontenc}
\usepackage{graphicx}
\usepackage{amssymb}
\usepackage{amsmath}
\begin{document}
\title{Simple Token-Efficient Vision-Language Model for Case-level Pathology Synoptic Report Generation}
\titlerunning{Token-Efficient VLM for Pathology Reports}
%
\author{Zhiyuan Yang\inst{1} $^{\ast}$ \and
Jiahao Cheng\inst{1} $^{\ast}$  \and
Vincent Quoc-Huy Trinh\inst{2, 3} \and
Mahdi S. Hosseini\inst{1, 4} \Envelope}
%
\authorrunning{Z. Yang et al.}
%
\institute{Department of Computer Science and Software Engineering (CSSE), Concordia University, Montreal, Canada \\ \email{mahdi.hosseini@concordia.ca}\\ \and
Axe Cancer, Centre de recherche du CHUM, Université de Montréal, Montreal, Canada \\ \and
Institut de recherche en immunologie et cancérologie (IRIC), Université de Montréal \\ \and
Mila - Quebec AI Institute, Montreal, Canada \\
$^{\ast}$ These authors contributed equally.}
\maketitle              
\begin{abstract}
Generating clinically useful pathology reports for pathology cases from whole-slide images (WSIs) is challenging due to gigapixel resolution, long visual-token sequences, and the complexity of case-level reasoning, where a single case may contain multiple WSIs with heterogeneous tissues and ambiguous findings. We present a simple token-efficient vision--language model for case-level synoptic report generation that remains practical under constrained GPU memory. Our architecture follows a minimal three-component design: a frozen pathology patch encoder, a lightweight two-layer MLP vision-language aligner, and a large language model decoder, with an explicit WSI marker token to separate slides within a case. Training proceeds in two supervised stages: (1) aligner-only WSI captioning using heterogeneous WSI-text pairs, and (2) case-level supervised fine-tuning on case-report pairs for structured report generation. To reduce sequence length, we represent each slide using $512 \times 512$ patches at $5\times$ magnification, which reduces the average sequence length by up to $64\times$ times compared to the commonly used $20\times$ patches. Combined with efficient training techniques, we enable practical training with only half a NVIDIA H100 GPU. Across both training stages, our approach achieves high ROUGE-L/METEOR/BLEU-4 scores while being substantially more efficient in memory and runtime. In AI-based evaluations, our model is consistently preferred over strong baselines. Extensive ablations characterize performance-efficiency trade-offs and identify simple choices that improve robustness in multi-WSI settings. Overall, this work provides a strong, reproducible baseline for efficient pathology report generation, lowering the barrier to multi-WSI VLM research under limited compute. Code is available at https://github.com/AtlasAnalyticsLab/PathoSynVLM.

\end{abstract}

\section{Introduction}
\begin{figure}[ht]
\centering
\includegraphics[width=\textwidth]{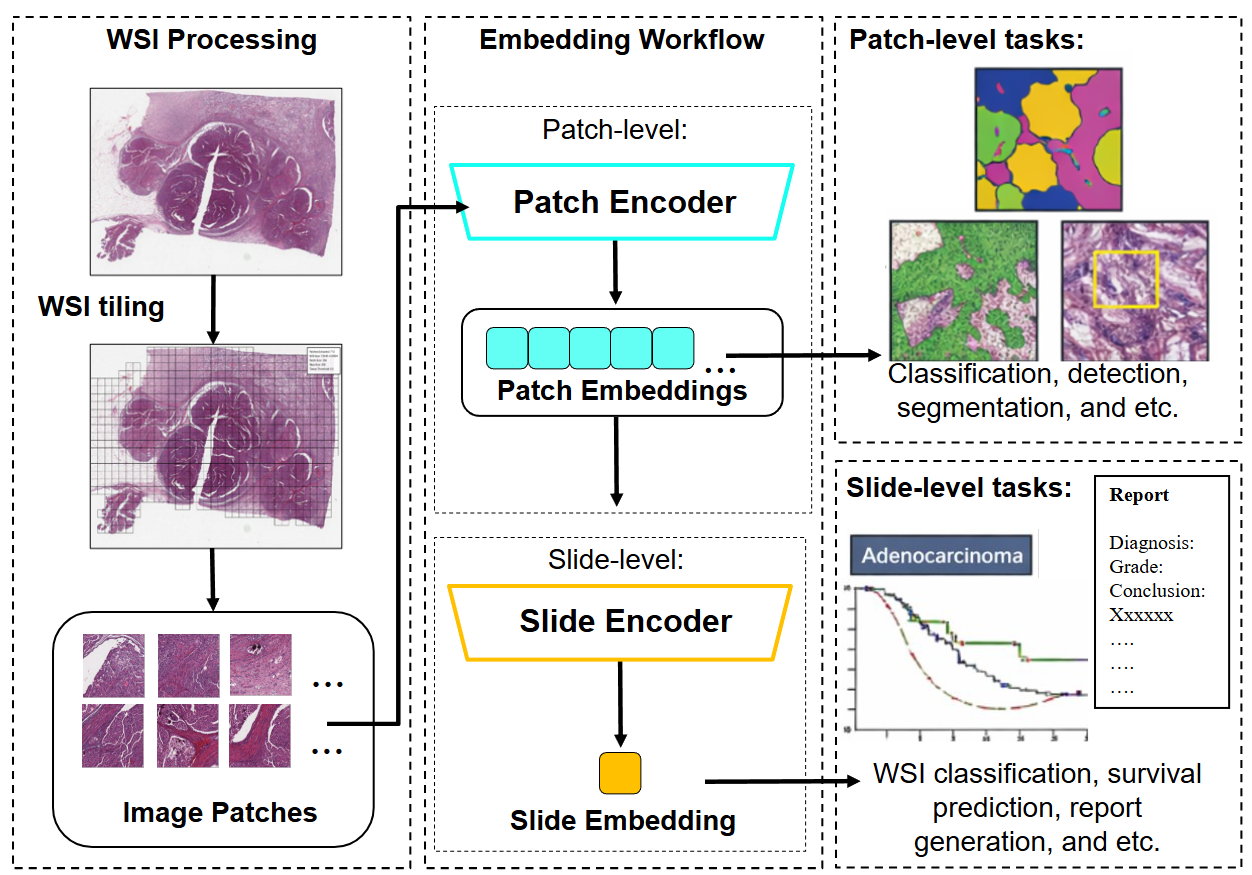}
\caption{Standard patch-based WSI processing workflow in computational pathology. Gigapixel WSIs are tiled into image patches, encoded by a patch encoder, and either used for patch-level prediction or aggregated into slide-level representations. The large number of patches extracted from each WSI creates a major memory and computation bottleneck for slide- and case-level modeling. }\label{workflow}
\end{figure}

In digital pathology, pathologists rely on whole-slide images (WSIs), which are gigapixel digital scans of histopathology slides, for disease diagnosis and the preparation of clinical pathology reports \cite{cpath}. With the development of computational pathology (CPath), deep-learning-based assisting tools have been developed for slide-level tasks like WSI classification, captioning and single-slide reporting \cite{cpath,titan,gigapath,Tran2025HistoGPT,prism}. However, WSIs pose a fundamental bottleneck for deep learning systems due to their extreme size, which typically requires patch-based processing that tiles a WSI into small image patches, as shown in Figure \ref{workflow}. Most commonly, image patches are extracted at high magnifications to capture details, which sometimes result in hundreds of thousands of patches being extracted per WSI and lead to extremely long sequences of input tokens to deep learning systems and cause efficiency/memory issues \cite{cpath,titan,gigapath,Tran2025HistoGPT,prism}. Furthermore, in practice, diagnostic decisions and reporting are made on a \emph{per-case} basis: a single patient case may include multiple tissue sections and staining protocols, resulting in multiple WSIs that must be jointly reviewed to reach a final interpretation \cite{cpath}. When transitioning from slide-level to case-level tasks, the WSI processing bottleneck becomes even more pronounced, as the model must handle multiple WSIs per case, increasing input length and memory cost during training while also raising the difficulty of coherent, evidence-consistent generation across slides \cite{pathttt,cpath}.

Recent slide-level pathology foundation models have introduced specialized mechanisms to cope with the extreme token lengths induced by WSIs, but these same design choices expose practical limitations when one moves to case-level report generation. Prov-GigaPath \cite{gigapath} serializes a WSI into a very long sequence of $256\times256$ patches at $20\times$ magnification and applies a dedicated slide encoder based on LongNet-style \cite{longnet} long-context modeling to integrate global context across the entire slide, which applies dilated/segmented attention that limits the model's ability to capture global context, and still relies on patch-heavy tokenization. PRISM \cite{prism} addresses the long-sequence issue by using a Perceiver-style \cite{blip2} latent bottleneck, where a fixed set of learned latent queries cross-attend to the patch embeddings to compress thousands of patch tokens into a compact representation. This adds an extra aggregation architecture and cross-attention stack on top of the patch encoder, increasing the overall model complexity and requires careful training. TITAN \cite{titan} handles the long-sequence issue during training using the train-short-test-long \cite{train-short} strategy by training only on local crops of WSIs instead of using the entire WSI, and relies on positional extrapolation to generalize to longer contexts during inference. This increases the risk of missing evidence during global aggregation if the evidence patches spread beyond the local crops. In common, these approaches all work on \textbf{$20\times$ high magnification patches}, and introduce \textbf{non-trivial architectural complexity} to manage extremely long visual sequences from WSI patch processing, yet none of them is explicitly designed for \emph{case-level} report generation where multiple WSIs per case must be aggregated under a strict memory budget. Furthermore, the training of these models often requires using \textbf{multiple high-end GPUs} (i.e. 4 to 32 NVIDIA A100 GPUs). Taken together, these foundations highlight a clear gap: we still lack a case-level VLM that can aggregate multiple WSIs efficiently within a tight memory budget, without relying on complex architectures or high-budget GPUs.

In this work, we address this gap by designing a \textbf{simple, token-efficient,} and \textbf{resource-aware} vision-language framework for \textbf{case-level} pathology report generation. Instead of introducing a specialized slide-level sequence model or latent-bottleneck aggregator, we couple strong frozen patch representations with a lightweight vision--language interface and a compact instruction-tuned large language model (LLM), and train the system using a straightforward two-stage supervised recipe. Furthermore, we address the long visual token sequence issue by using low-magnification image patches instead of the standard high-magnification patches. This yields a practical baseline that explicitly accounts for multi-WSI cases and can be trained under constrained GPU memory budgets while retaining competitive report-generation quality. Specifically, this study has the following contributions:

\begin{itemize}
    \item 
    We propose a deliberately \textbf{lightweight case-level VLM} that avoids specialized long-context slide encoders and latent-bottleneck aggregators. The model consists of a frozen pathology patch encoder, a two-layer MLP vision-language aligner, and a LLM decoder. To support multi-WSI cases, we introduce an explicit \textbf{WSI marker token} to indicate slide boundaries, enabling a single report to be generated from a variable number of WSIs within a fixed memory budget.

    \item 
    We develop a \textbf{straightforward training pipeline} that separates vision-language alignment from case-level structured generation. In stage~1, we train only the aligner using WSI captioning on a heterogeneous mixture of WSI--text datasets, establishing a robust interface between frozen visual embeddings and the LLM. In stage~2, we perform supervised fine-tuning with teacher forcing on a case-report dataset to produce schema-consistent synoptic outputs, showing that a minimal interface plus standard supervision can already yield practical case-level report generation behavior.

    \item 
    To explicitly address the extreme visual-token lengths of WSIs, we study a \textbf{low-magnification patching strategy} that uses $512\times512$ patches at $5\times$ magnification instead of the commonly adopted $20\times$ regimes. We show that high-quality report generation across a variety of organs remains feasible under this token budget, while reducing the average number of input patches by roughly an order of magnitude. We further quantify the performance-efficiency trade-off when compressing the visual sequence more aggressively (down to $1\times$), providing concrete guidance on where quality begins to degrade.

    \item 
    We demonstrate that training a Qwen2.5-3B-based \cite{qwen2025qwen25technicalreport} pathology VLM is feasible on \textbf{half of an NVIDIA H100 GPU} by combining efficiency training techniques, including LoRA \cite{lora}, gradient accumulation, activation checkpointing, and \texttt{bf16} mixed-precision training. This offers a reproducible baseline for groups operating under constrained compute.
\end{itemize}
\section{Methodology}
\label{sec:methodology}
In this section, we describe a simple resource-aware pathology VLM for case-level report generation. The method consists of a frozen patch encoder, a two-layer MLP vision-language aligner, and an LLM decoder trained with a two-stage supervised recipe. We also describe the efficiency techniques that make the pipeline feasible under limited GPU memory.

\subsection{Model Architecture}

\begin{figure}[ht]
\centering
\includegraphics[width=\textwidth]{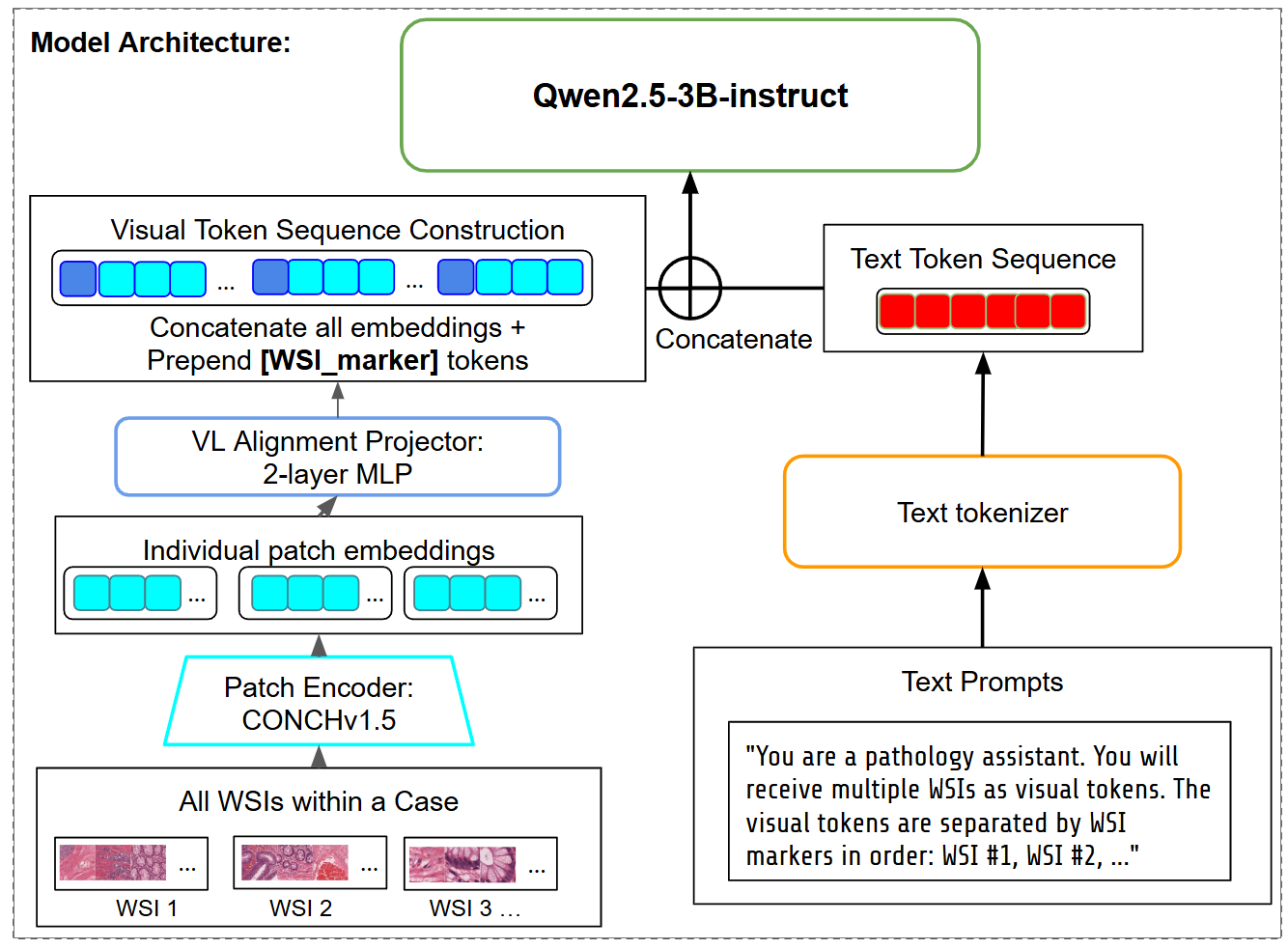}
\caption{Our model follows the LLaVA-like \cite{llava} design with three major components: 1) a frozen patch-level encoder that embeds individual WSI patches into visual feature embeddings, 2) a two-layer MLP that projects visual feature embeddings to the language token embedding space, and 3) a LLM that takes in a sequence of concatenated visual token embeddings with text tokens for report generation. When constructing the visual token sequence, we prepend a learnable [WSI\_marker] token before each WSI subsequence to separate WSIs within a case. The visual token sequence is concatenated with tokenized text prompts as a prefix. The entire multimodal sequence is then used as input to the LLM for report generation. } \label{fig1}
\end{figure}

Our model follows the LLaVA-like \cite{llava} design and is intentionally designed to be simple, modular, and compatible with limited computational resources while still supporting case-level pathology report generation. As illustrated in Figure \ref{fig1}, the architecture contains three components: (i) a frozen patch-level visual encoder that maps image patches to continuous embeddings, (ii) a lightweight two-layer MLP that aligns visual embeddings to the language model embedding space, and (iii) a LLM that autoregressively generates text conditioned on the aligned visual tokens. Specifically, WSIs are first processed independently by tissue segmentation and patch sampling to extract a set of tissue patches from each slide. Each patch is then passed through a frozen patch-level encoder to obtain a sequence of patch embeddings per WSI, which are mapped into the language-model embedding space by a two-layer MLP vision--language aligner. For a case containing multiple WSIs, the aligned patch-token sequences are concatenated into a single case-level visual sequence, with a learnable WSI marker token inserted between successive WSIs to explicitly indicate slide boundaries. In parallel, a text input sequence is constructed using a training-stage-specific instruction prompt, tokenized by the LLM tokenizer and converted to text token embeddings by the LLM embedding layer. Finally, the model forms one multimodal sequence by concatenating the aligned visual prefix and the text embeddings, and feeds it into the LLM decoder for autoregressive generation of the case-level report conditioned on the visual evidence from all WSIs. In the following paragraphs, we explain the components of our architecture and workflow in detail.

\textbf{Patch-level visual encoding.} We adopt CONCHv1.5 \cite{lu2024conch} as the patch-level visual encoder. CONCH is a pathology-specific foundation model that produces semantically rich embeddings for histopathology image patches, trained to capture morphological patterns and stain-related variations that are difficult to learn from scratch with limited labeled data. Using a strong domain-aligned patch encoder provides a robust visual representation for downstream vision--language training, and allows the language model to focus on translating these representations into clinically meaningful textual descriptions rather than relearning low-level visual features.

\textbf{Vision-language alignment via a two-layer MLP.} The core trainable vision-side module is a two-layer MLP projector with a GeLU \cite{Gelu} activation function in between, which maps patch embeddings into the LLM hidden dimension. We intentionally choose this small projector instead of a fusion transformer, Perceiver module, or cross-attention stack for three reasons. First, the visual features are produced by a frozen pathology-pretrained encoder, so the projector does not need to learn low-level histology representations from scratch; its main role is to translate already meaningful visual embeddings into the LLM input space. Second, case-level pathology inputs can contain many WSIs, making the visual sequence length the dominant memory bottleneck; adding another trainable visual aggregation module would increase both architectural complexity and activation memory. Third, using a minimal projector makes the system easier to reproduce and provides a clearer baseline for studying whether simple alignment plus LLM self-attention can support structured case-level generation. This choice is not intended to imply that MLP alignment is universally optimal; rather, it reflects our goal of establishing a lightweight baseline before introducing heavier multimodal fusion mechanisms.

\textbf{Prefix conditioning and multimodal sequence construction.} After alignment, the projected visual tokens are treated as a prefix to the text input of the LLM. Concretely, we construct a single multimodal sequence by concatenating the aligned visual token sequence with the prompt and any textual context tokens, and then perform standard causal decoding. This prefix conditioning design keeps the fusion mechanism minimal and leverages the LLM's native self-attention to integrate visual evidence while generating text. During supervised training, we mask the loss over the visual prefix so that the language modeling loss is computed only over text tokens, ensuring that the model learns to predict the desired report text conditioned on the visual content.

\textbf{WSI boundary modeling with a learnable marker token.} Case-level inputs may contain multiple WSIs, and their ordering is not necessarily informative. To explicitly represent slide boundaries within a single packed visual sequence, we introduce a learnable \emph{WSI marker} token that is inserted between the patch-token groups of successive WSIs. This marker serves two purposes: (1) it provides a clear separation cue so the LLM can avoid blending evidence from different slides, and (2) it enables the model to learn a consistent internal structure for multi-WSI cases while still operating on a single concatenated sequence. We activate this mechanism only in the second training stage, where the supervision targets are case-level structured reports with multi-WSI inputs.

\textbf{LLM backbone and stage-specific prompts.} We use Qwen2.5-3B-Instruct \cite{qwen2025qwen25technicalreport} as the language backbone. Since we train our model in two stages, we use two stage-specific instruction prompts to align the model behavior with the supervision target in each stage without changing the architecture. In stage~1, the prompt requests a concise summary of observations from the input WSI, matching the caption-style supervision used for vision-language alignment. In stage~2, the prompt instructs the model to produce a case-level report following a fixed structured format with multiple WSIs separated by the WSI marker tokens. This prompt-based task specification provides a simple and effective control mechanism and reduces ambiguity in the desired completion format. Details of the format requirement and prompts are provided in Section \ref{subsec:setup}.

\subsection{Training Recipe}
\begin{figure}[ht]
\includegraphics[width=\textwidth]{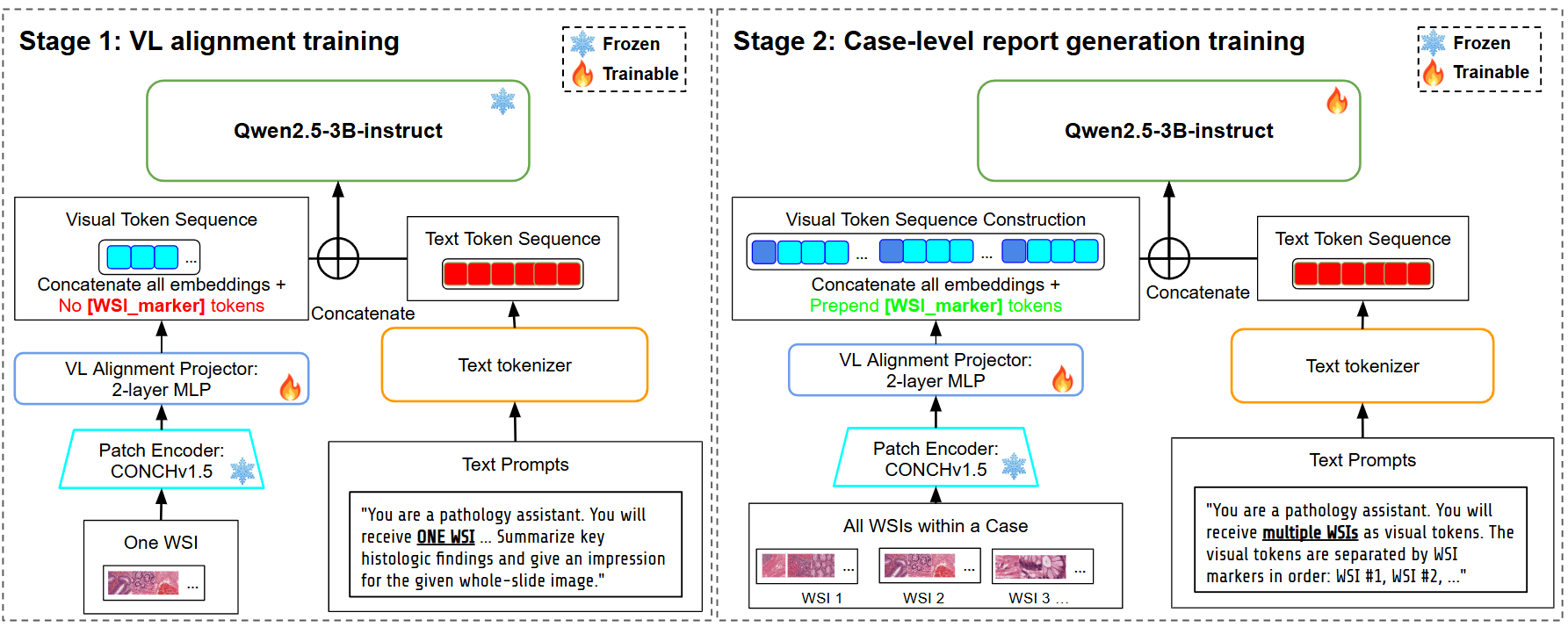}
\caption{We have two training stages. In stage 1, everything is frozen, except for the two-layer MLP for vision-language alignment. We train via the WSI captioning task on individual WSIs, which requires the model to understand the visual tokens from each WSI and generate a summary/caption for the given WSI.  In stage 2, we only freeze the patch encoder and finetune both the two-layer MLP and the LLM backbone. An additional learnable \texttt{[WSI\_marker]} token is inserted before each WSI. We perform supervised finetuning using case-report pairs where the model is given a concatenation of multiple WSIs within the case, and asked to generate a report for the entire case. In both stages, we input task/stage-specific prompts to guide the model for the corresponding generation task. We introduce the details of the prompts and format requirements in Section \ref{subsec:setup}.} \label{training}
\end{figure}

As illustrated in Figure \ref{training}, we adopt a simple two-stage supervised training procedure designed to (i) first establish stable vision--language alignment with minimal trainable parameters, and (ii) then adapt the aligned model to case-level structured report generation. Both stages use standard autoregressive next-token prediction with teacher forcing, differing only in which parameters are updated and in the supervision data, and the prompts used to guide the model for the corresponding tasks. Details of the datasets used in both training stages are provided in Section \ref{sec:datasets}.

\textbf{Stage~1: vision--language alignment via supervised WSI captioning.} In the first stage, we freeze the patch encoder and the LLM, and train only the two-layer MLP alignment module. The model is optimized on a WSI captioning objective using paired WSI--caption data from a mixture of datasets. Given a WSI represented as a sequence of patch embeddings, the aligner projects these visual tokens into the LLM embedding space and provides them as a visual prefix to the LLM. An instruction prompt then asks the LLM to summarize salient observations from the input WSI, and the model is trained to predict the ground-truth caption tokens under teacher forcing via the standard cross-entropy loss. Restricting learning to the alignment module in this stage reduces the risk of overfitting the language model to dataset-specific caption styles and provides a controlled setting to learn a robust mapping from pathology visual features to the LLM input space. 

\textbf{Stage~2: supervised fine-tuning for case-level report generation.} In the second stage, we adapt the model to the downstream task of generating case-level structured pathology reports using a case-report dataset. In contrast to stage~1, we jointly train the MLP aligner and fine-tune the LLM. Each training example consists of one case potentially containing multiple WSIs; their patch-token sequences are packed into a single case-level visual prefix with explicit WSI boundary markers, and a stage-specific prompt instructs the model to produce a report following a predefined format. We again apply teacher forcing and optimize the next-token prediction (cross-entropy) loss on the target report text. This stage enables the model to learn task-specific phrasing, formatting, and domain conventions, while retaining the vision--language grounding established during the alignment stage. We introduce the details of the prompts and format requirements in Section \ref{subsec:setup}.

\subsection{Efficiency Techniques}

A key goal of our approach is to make case-level pathology VLM training feasible under limited GPU memory while keeping the model and recipe easy to reproduce. We therefore combine several complementary efficiency techniques that reduce both \emph{input-token} cost and \emph{trainable-parameter} cost, enabling training with a Qwen2.5-3B backbone on a single 40 GB GPU.

\textbf{Token-efficient visual inputs.} WSI-based models are often bottlenecked by the number of visual tokens rather than model size alone. Most of the existing works tile WSI into small patches ($256\times256$ or $224\times224$) at high magnification ($20\times$ and $10\times$) \cite{titan,gigapath,prism}, leading to hundreds of thousands of image patches per WSI. To control this cost, we represent each WSI using a small set of large-field-of-view patches: tissue patches of size $512\times512$ extracted at $5\times$ magnification (equivalent to $2.5\times$ when resized to $256\times256$ before patch embedding), which substantially reduces the length of the visual token sequence by $64\times$ times compared to common high-magnification dense patching schemes (e.g., $256\times256$ patches at $20\times$). We further evaluate an extreme low-token setting using $512\times512$ patches extracted at only $1\times$ magnification in the ablative studies to characterize the accuracy--efficiency trade-off. The choice of 5× magnification reflects a trade-off between morphological detail and sequence length. Dense 20× patching preserves cytologic detail but produces very long token sequences, which becomes especially restrictive when multiple WSIs must be packed into a single case-level input. In contrast, 1× patching is highly efficient but primarily captures coarse tissue layout and loses histologic information needed for reliable report generation. We therefore use 5× as a mesoscopic compromise: it preserves broader tissue architecture, lesion distribution, and specimen-level context while keeping the token budget compatible with low-memory multi-WSI training. This choice is later supported by our magnification ablation, where 1× substantially reduces generation metrics, and by the efficiency comparison showing the high cost of 20× patching.


\textbf{Precomputed patch embeddings with a frozen visual encoder.} To avoid backpropagating through the vision backbone and to decouple WSI processing from VLM training, we precompute patch embeddings offline using a frozen CONCHv1.5 encoder \cite{lu2024conch}. During training, the VLM consumes these embeddings directly as its visual input, which significantly reduces GPU memory and accelerates experimentation without repeatedly running the patch encoder.

\textbf{Parameter-efficient fine-tuning of the LLM.} In stage~2, we fine-tune the language backbone using Low-Rank Adaptation (LoRA) \cite{lora}, which injects a small number of trainable low-rank matrices into selected linear projections while keeping the original LLM weights frozen. In our setting, we only finetune the query and key projection layers within the attention blocks. This substantially reduces the number of trainable parameters and the optimizer state, thereby lowering memory usage under limited compute resources. We jointly update the MLP aligner alongside LoRA parameters to adapt both the vision-language interface and the generation behavior to the structured case-level reporting task.

\textbf{Memory-saving training strategies.} We additionally employ three techniques to fit training into constrained VRAM budgets. First, we use gradient accumulation to simulate larger effective batch sizes without increasing instantaneous memory consumption. Second, we use gradient checkpointing to trade additional compute for reduced activation memory by recomputing intermediate activations during backpropagation. Third, we train with mixed precision using automatic mixed precision in \texttt{bf16} precision, which further lowers memory footprint and improves throughput while preserving numerical stability on modern GPUs. Together, these choices enable end-to-end VLM training with the Qwen2.5-3B backbone on half of an NVIDIA H100 GPU (40 GB VRAM), while maintaining the same optimization objective and overall training workflow.
\section{Dataset}
\label{sec:datasets}
\begin{figure}[ht]
\centering
\includegraphics[width=\linewidth]{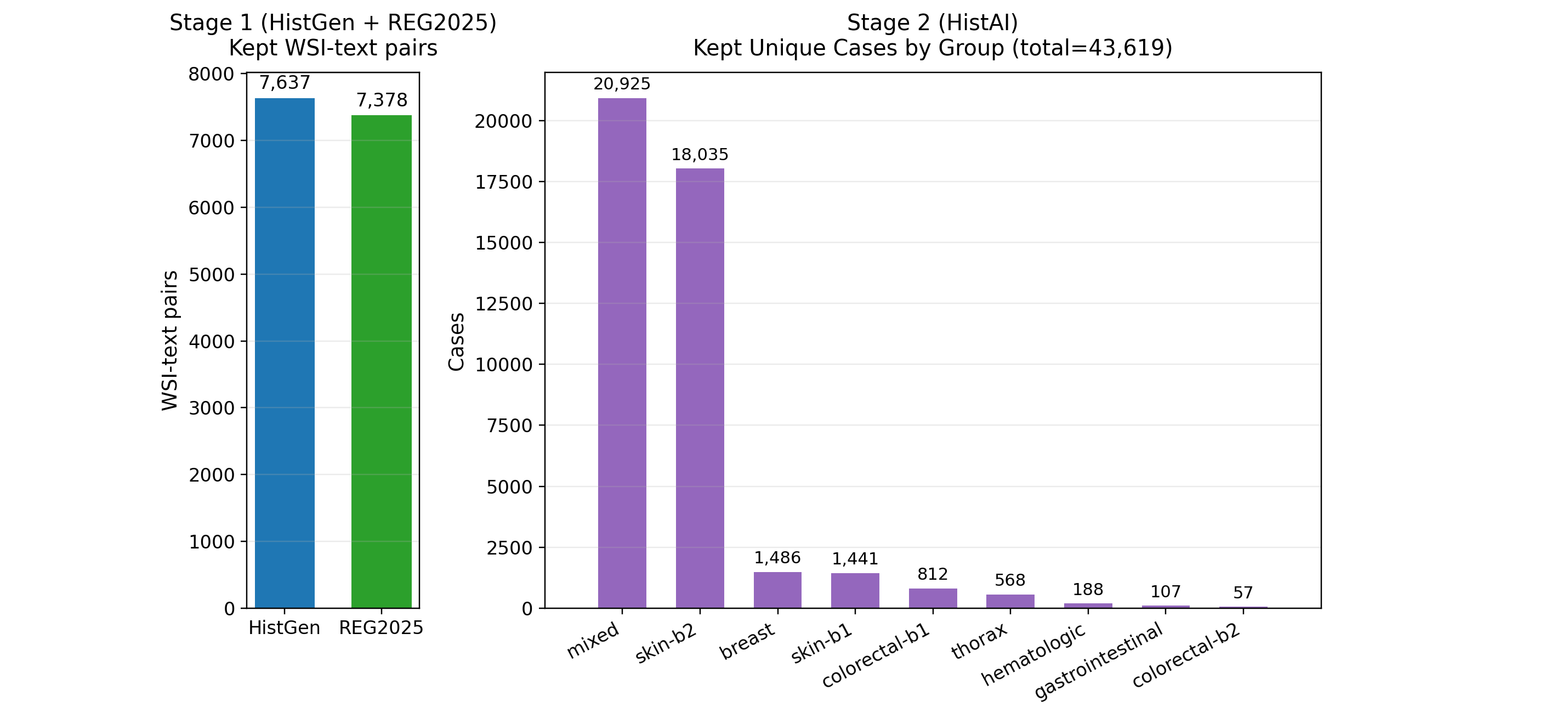}
\caption{\textbf{Dataset composition used in this work.} The left panel shows retained WSI-text pairs in Stage~1. The right panel shows retained HISTAI cases per domain in Stage~2. Stage~2 is imbalanced, with skin and mixed-domain cases dominating the dataset.}
\label{fig:dataset_distribution}
\end{figure}

We use three datasets across the two training stages: REG2025 \cite{REG2025}, HistGen \cite{Guo_HistGen_MICCAI2024}, and HISTAI \cite{nechaev2025histaiopensourcelargescaleslide}. Figure~\ref{fig:dataset_distribution} summarizes the retained WSI-text pairs for Stage~1 and the retained HISTAI case-report pairs for Stage~2 after standardization, feature discovery, and integrity checks.

\textbf{Data preprocessing on Stage~1.} Stage~1 trains the aligner with paired WSI-to-text supervision from HistGen and REG2025 \cite{Guo_HistGen_MICCAI2024,REG2025}. HistGen provides TCGA-derived WSIs paired with diagnostic report text \cite{Tomczak2015TheCG}, while REG2025 \cite{REG2025} provides WSIs matched to CAP-style structured reports across seven organs and multiple institutions. All WSIs are processed with AtlasPatch \cite{alagha2026atlaspatch} for tissue detection, coordinate extraction, and CONCHv1.5 feature embedding \cite{lu2024conch}. We use $512\times512$ tissue patches at $5\times$ magnification by default and include a $1\times$ configuration for the magnification ablation. Target texts are standardized into \texttt{image\_id} and \texttt{report} fields, and samples with missing text, missing features, or unreadable files are removed. Table~\ref{tab:stage1_accounting} reports the retained Stage~1 statistics.

\begin{table}[h]
\caption{\textbf{Stage~1 dataset accounting after metadata standardization and integrity filtering.} For each supervision source, we report retained WSI-text pairs, patch counts at $5\times$ and $1\times$ magnification using $512\times512$ patches, and average patches per WSI.}
\label{tab:stage1_accounting}
\centering
\scriptsize
\setlength{\tabcolsep}{4pt}
\begin{tabular}{|l|r|r|r|r|r|}
\hline
\multicolumn{2}{|c|}{} &
\multicolumn{2}{c|}{$5\times$ magnification, $512\times512$ patches} &
\multicolumn{2}{c|}{$1\times$ magnification, $512\times512$ patches} \\
\hline
Dataset & \# Pairs & \# Patches & \# Avg patches/WSI & \# Patches & \# Avg patches/WSI \\
\hline
HistGen & 7\,637 & 1\,821\,870 & 238.56 & 96\,938 & 12.69 \\
REG2025 & 7\,378 & 581\,454 & 78.81 & 45\,290 & 6.14 \\
\hline
\end{tabular}
\end{table}

\textbf{Data preprocessing on Stage~2.} Stage~2 uses HISTAI \cite{nechaev2025histaiopensourcelargescaleslide} for case-level supervised fine-tuning. HISTAI contains 112{,}801 WSIs from 47{,}279 cases with metadata fields including diagnosis and conclusion. We apply the same preprocessing and integrity filtering protocol as in Stage~1, but only use $512\times512$ tissue patches at $5\times$ magnification. After filtering, the retained cohort contains 43{,}618 cases, 102{,}703 WSIs, and 16{,}331{,}401 tissue patches across nine subsets. Table~\ref{tab:stage2_histai_distribution} reports the per-domain statistics.

\begin{table}[h]
\caption{\textbf{Stage~2 HISTAI dataset statistics after preprocessing.} For each subset, we report the number of retained cases, WSIs, tissue patches, and averages of WSIs per case and patches per WSI/case.}
\label{tab:stage2_histai_distribution}
\centering
\scriptsize
\resizebox{\linewidth}{!}{
\begin{tabular}{|l|r|r|r|r|r|r|}
\hline
Group & \# Used Cases & \# WSIs & \# Patches & \# Avg WSIs/Case & \# Avg Patches/WSI & \# Avg Patches/Case \\
\hline
HISTAI-breast & 1\,486 & 1\,679 & 345\,382 & 1.130 & 205.71 & 232.42 \\
HISTAI-colorectal-b1 & 812 & 4\,436 & 826\,795 & 5.463 & 186.38 & 1\,018.22 \\
HISTAI-colorectal-b2 & 57 & 88 & 15\,241 & 1.544 & 173.19 & 267.39 \\
HISTAI-gastrointestinal & 107 & 183 & 28\,954 & 1.710 & 158.22 & 270.60 \\
HISTAI-hematologic & 188 & 188 & 40\,849 & 1.000 & 217.28 & 217.28 \\
HISTAI-mixed & 20\,924 & 52\,019 & 8\,271\,097 & 2.486 & 159.00 & 395.29 \\
HISTAI-skin-b1 & 1\,441 & 6\,206 & 1\,282\,142 & 4.307 & 206.60 & 889.76 \\
HISTAI-skin-b2 & 18\,035 & 37\,173 & 5\,390\,889 & 2.061 & 145.02 & 298.91 \\
HISTAI-thorax & 568 & 731 & 130\,052 & 1.287 & 177.91 & 228.96 \\
\hline
Total/Avg & 43\,618 & 102\,703 & 16\,331\,401 & 2.355 & 159.02 & 374.42 \\
\hline
\end{tabular}
}
\end{table}


\section{Experiments and Results}
\label{sec:experiments_results}
We evaluates the proposed two-stage training strategy in terms of generation quality, efficiency, and clinical relevance. Stage~1 evaluates single WSI captioning with ROUGE-L \cite{lin-2004-rouge}, METEOR \cite{lavie-agarwal-2007-meteor}, BLEU-4 \cite{papineni-etal-2002-bleu}, and BERTScore-F1 \cite{Zhang2020BERTScore}. Efficiency is evaluated by reporting token usage and runtime. Stage~2 requires a fixed format check with explicit fields, and results are reported at both the whole-output and field levels. Because lexical metrics are imperfect proxies for clinical correctness, we additionally report AI-assisted comparisons and a targeted pathologist audit.

\subsection{Experiment Setup}
\label{subsec:setup}

We summarizes the experimental configuration shared across both stages, including the unified metadata protocol, prompting and target formats, evaluation metrics, comparison strategies and optimization settings. All experiments are performed on the test split of the dataset used in both stages.

\textbf{Format.}
A unified text format schema standardizes supervision across datasets and stages and facilitates evaluation. Each sample is serialized in a compact \texttt{json} format that stores a unique field identifier and the corresponding target text, allowing for consistent data loading. In Stage~1, targets are free-form captions or reports with no specific format constraints. In Stage~2, the model is trained to generate structured reports with explicit field headers: \texttt{Diagnosis}, \texttt{Certainty}, and \texttt{Conclusion}. This structure reduces format ambiguity and enables field-level evaluation aligned with reporting requirements.

\begin{figure}[htp]
\centering
\includegraphics[width=0.8\linewidth]{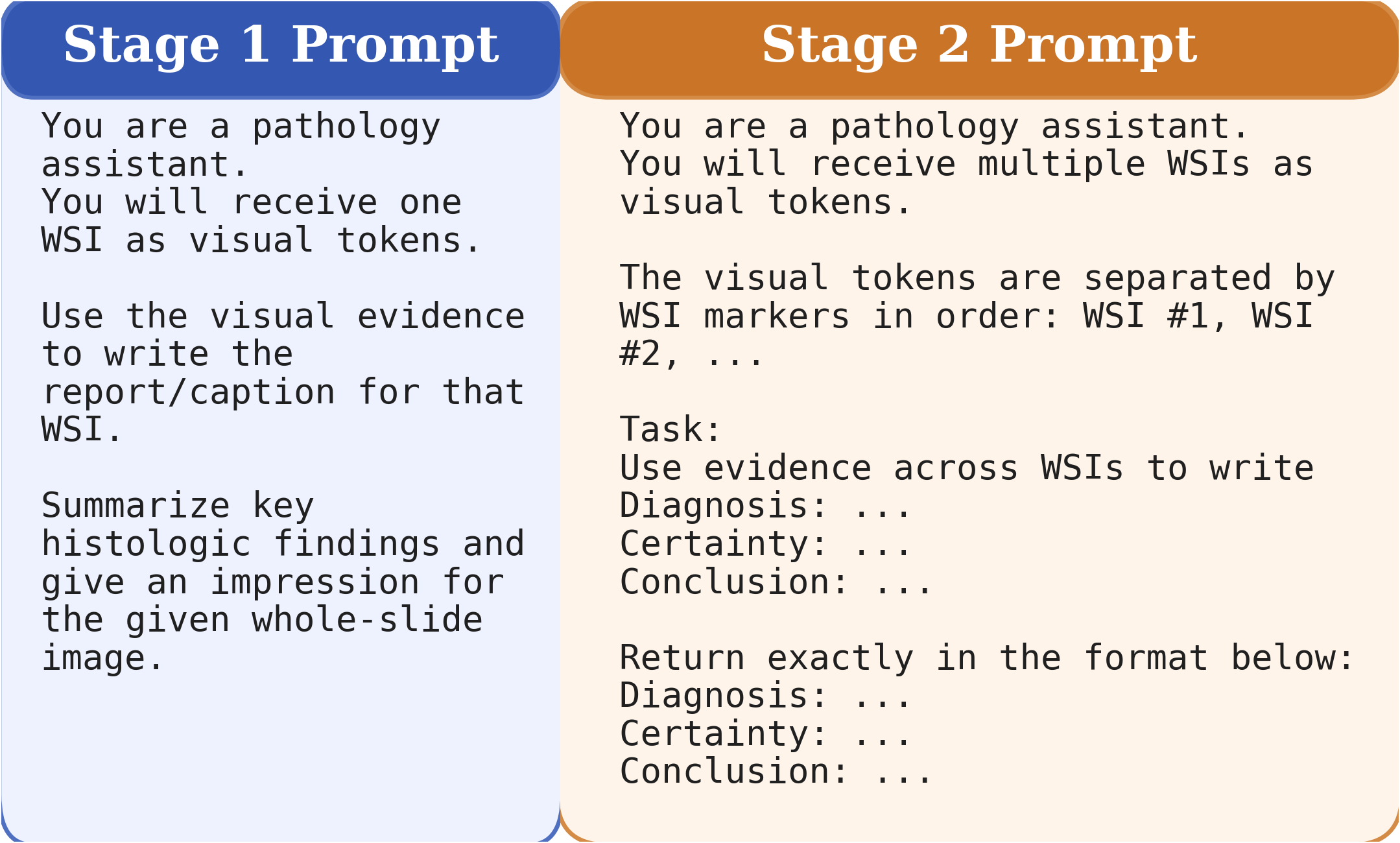}
\caption{\textbf{Prompts used in the two-stage training pipeline.} Left shows the Stage~1 single-WSI prompt, which conditions the model on visual tokens from one WSI and requests a free-form slide-level caption or report describing key histologic findings and an overall impression. Right shows the Stage~2 multi-WSI prompt, which packs multiple WSIs in a fixed order separated by WSI markers and requires case-level structured generation with explicit fields for Diagnosis, Certainty, and Conclusion.}
\label{fig:prompts_stage12}
\end{figure}

\textbf{Prompt.}
Prompt design has been shown to play an important role in guiding the model in different tasks \cite{prompt}. In this work, prompts are constructed to match the supervision granularity of each stage and to make the input structure explicit. Stage~1 employs a single-WSI instruction prompt, as generation is conditioned on one slide at a time for caption or report generation. Stage~2 uses case-level prompts that may include multiple WSIs, with explicit marks of WSI boundaries indicated by \texttt{WSI \#1}, \texttt{WSI \#2}, and so on. This approach guides the model to aggregate evidence across slides while producing the required structured output. Figure \ref{fig:prompts_stage12} shows the prompts used in both stages.

\textbf{Evaluation Metrics.}
Stage~1 is evaluated using ROUGE-L \cite{lin-2004-rouge}, METEOR \cite{lavie-agarwal-2007-meteor}, BLEU-4 \cite{papineni-etal-2002-bleu}, and BERTScore-F1 \cite{Zhang2020BERTScore}. Stage~2 uses these metrics on the \texttt{Conclusion} field and additionally reports field-level correctness. Certainty match is the exact match rate of the normalized \texttt{Certainty} field. Diagnosis-relaxed match uses a containment criterion after normalization, where a prediction is counted as correct if the reference diagnosis contains the prediction or the prediction contains the reference. These automatic metrics are useful for reproducibility but are not treated as clinical validation, motivating the expert audit described below.

\textbf{AI-based comparison protocol.}
We use ChatGPT 5.4 \cite{openai_chatgpt_2026} as the judge to compare our Stage~2 outputs with HistoGPT \cite{Tran2025HistoGPT}, PRISM \cite{prism}, and WSI-LLaVA \cite{wsi-llava}. Because these models follow different reporting conventions, comparisons use the largest compatible reference target for each model. PRISM \cite{prism} generally produces short statements that only produce diagnosis, so its comparison is restricted to diagnosis content. HistoGPT \cite{Tran2025HistoGPT} and WSI-LLaVA \cite{wsi-llava} often generate microscopic descriptions, so we use the HISTAI \cite{nechaev2025histaiopensourcelargescaleslide} \texttt{micro protocol} field for comparison. For each valid case, the judge assigns a score from $0$ to $10$ and selects the preferred output. We interpret these results as a complementary comparison rather than a direct measure of diagnostic correctness.

\textbf{Pathologist audit.}
We conducted a targeted expert audit on the overlapping skin-case subset used for the baseline vs. HistoGPT \cite{Tran2025HistoGPT} comparison. A senior certified pathologist reviewed $67$ cases containing the baseline output, the HistoGPT output, the clinical diagnosis field, the microscopic protocol, and the reference conclusion. The audit focused on the reference conclusion and final diagnosis because the clinical diagnosis field reflects information provided to the pathologist before slide review and is therefore not an appropriate prediction target. In addition, our baseline does not generate a microscopic protocol section, so microscopic description scoring was not directly comparable across models. Each output was scored using diagnostic consistency with the reference conclusion (0--3) and coverage of key conclusion-level findings (0--2), giving a composite score from $0$ to $5$.

\textbf{Efficiency Measurement.}
To contextualize computational cost, we evaluate efficiency under the default $5\times$ patching setting and a higher-resolution $20\times$ setting while keeping all other components fixed. For each setting, we report the number of extracted tissue patches per input, the preprocessing time, and the inference time. 

\textbf{Parameter Configurations.}
All experiments follow a consistent optimization and decoding configuration under computational constraints. Training uses batch size 1 with gradient accumulation 8 and the AdamW optimizer, with learning rate $10^{-4}$, weight decay $0.01$, and a cosine learning rate schedule with warmup ratio $0.03$. All runs use bf16 precision to reduce memory usage and improve throughput while maintaining numerical stability during large scale vision language training.

\subsection{Quantitative Results}
\label{subsec:quant_results}

We report model performance and efficiency for both stages. Performance is measured with standard text-generation metrics. In stage~2, we additionally report field-level correctness under the required structured schema. Efficiency is summarized by the average number of tissue patches processed per input and the per-input runtime for offline preprocessing and report-generation inference.

\begin{table}[ht]
\caption{\textbf{Quantitative performance for both stages.} Stage~1 evaluates slide-level free-form generation under single WSI inputs. Stage~2 evaluates case-level structured reporting, and overlap metrics are computed on the Conclusion field. ROUGE-L, METEOR, BLEU-4, and BERTScore-F1 measure similarity between generated text and reference targets. Diagnosis exact match, diagnosis relaxed match, and certainty exact match measure field-level correctness for Stage~2.}
\label{tab:perf_stage12}
\centering
\small
\setlength{\tabcolsep}{2.2pt}
\renewcommand{\arraystretch}{1.12}
\begin{tabular}{@{}l c c c c c c c@{}}
\hline
Setting & \multicolumn{4}{c}{Text-generation metrics} & \multicolumn{3}{c}{Field-level correctness} \\
\cline{2-5}\cline{6-8}
 & ROUGE-L & METEOR & BLEU-4 & \shortstack{BERTScore\\F1} & \shortstack{Diag.\\exact} & \shortstack{Diag.\\relaxed} & \shortstack{Cert.\\exact} \\
\hline
Stage~1 baseline & 0.4743 & 0.4810 & 0.1247 & 0.4253 & N/A & N/A & N/A \\
Stage~2 baseline & 0.2495 & 0.1988 & 0.0525 & 0.3018 & 0.1667 & 0.3333 & 0.9000 \\
\hline
\end{tabular}
\end{table}

\begin{table}[ht]
\caption{\textbf{Stage~1 efficiency under single WSI inputs.} Patches per input reports the average number of tissue patches processed per WSI. Preprocessing seconds per input measures the time spent on patch extraction and embedding generation per WSI. Inference seconds per input measures the report generation runtime per WSI. The $20\times$ magnification setting uses the same pipeline with a higher-resolution patching configuration.}
\label{tab:stage1_eff_only}
\centering
\small
\setlength{\tabcolsep}{5pt}
\renewcommand{\arraystretch}{1.15}
\begin{tabular}{@{}l c c c@{}}
\hline
Setting &
\shortstack{Patches\\per input} &
\shortstack{Preprocessing seconds\\per input} &
\shortstack{Inference seconds\\per input} \\
\hline
Baseline & 160.04 & 45.98 & 1.87 \\
$20\times$ baseline & 12,247.74 & 196.84 & 4.62 \\
\hline
\end{tabular}
\end{table}

For stage~1, we observe an ROUGE-L of $0.4743$ and METEOR of $0.4810$. The baseline configuration processes an average of $160.04$ patches per WSI and requires $45.98$ seconds of preprocessing per input. Inference requires $1.87$ seconds per input. When increasing the patch magnification to $20\times$, the average patch count rises substantially to $12{,}247.74$, resulting in a preprocessing time of $196.84$ seconds per input and an inference time of $4.62$. These findings indicate that the overall runtime is dominated by the number of extracted patches and the cost of embedding generation, highlighting a trade-off between information density and efficiency.

In stage~2, text overlap metrics are lower with ROUGE-L $0.2495$ and METEOR $0.1988$. It is expected because stage~2 performs case-level structured reporting with multi-WSI aggregation and strict output constraints, making the task both longer-horizon and less amenable to surface-form overlap. Diagnosis matching remains more challenging, with an exact match rate of $0.1667$ and a relaxed match rate of $0.3333$, reflecting the lexical diversity of diagnosis strings and the brittleness of exact matching under free-form phrasing. In terms of efficiency, our pipeline processes $374.42$ patches per case and requires $14.06$ seconds of preprocessing, while inference takes $2.4$ seconds per case. The increased inference cost is consistent with case-level generation and longer structured outputs. Efficiency analysis for stage~2 is not included because the stage~1 analysis already identifies the dominant cost factors. 

\begin{table}[ht]
\centering
\caption{\textbf{Pathologist audit on 67 overlapping skin cases.} Diagnostic consistency is scored from 0 to 3 against the reference conclusion/final diagnosis. Coverage is scored from 0 to 2 for key conclusion-level findings.}
\label{tab:pathologist_audit}
\small
\setlength{\tabcolsep}{4pt}
\begin{tabular}{lcc}
\hline
Metric & Baseline & HistoGPT \\
\hline
Diagnostic consistency, mean /3 & 0.85 & 0.61 \\
Coverage of key findings, mean /2 & 1.64 & 1.63 \\
Composite score, mean /5 & 2.49 & 2.24 \\
Exact or clinically equivalent diagnosis & 12/67 & 5/67 \\
Partially or fully consistent diagnosis ($\geq 2$) & 18/67 & 12/67 \\
Covers most key findings & 47/67 & 44/67 \\
Pairwise composite wins & 20 & 8 \\
Ties & 39 & 39 \\
\hline
\end{tabular}
\end{table}

The pathologist audit provides a clinically grounded complement to lexical metrics and AI-based preference judging. On the common conclusion-level target, the Stage 2 baseline achieves a higher mean diagnostic-consistency score than HistoGPT \cite{Tran2025HistoGPT} (0.85 vs. 0.61 out of 3) and a modestly higher composite score (2.49 vs. 2.24 out of 5). Pairwise comparison also favours the baseline more often than HistoGPT, although most cases are ties. Importantly, the absolute diagnostic-consistency scores are low for both systems, indicating that neither model should be interpreted as clinically adequate. These results suggest that fluent microscopic prose, coverage of general findings, and final-diagnosis correctness should be evaluated separately.

\subsection{Qualitative Results}
\label{subsec:qual_results}

To better characterize model behaviour beyond automatic metrics, we examine representative validation examples. The pathologist audit in Table~\ref{tab:pathologist_audit} complements these examples by assessing conclusion-level reliability on the overlapping skin-case subset.

\begin{figure}[ht!]
\centering
\includegraphics[width=\linewidth]{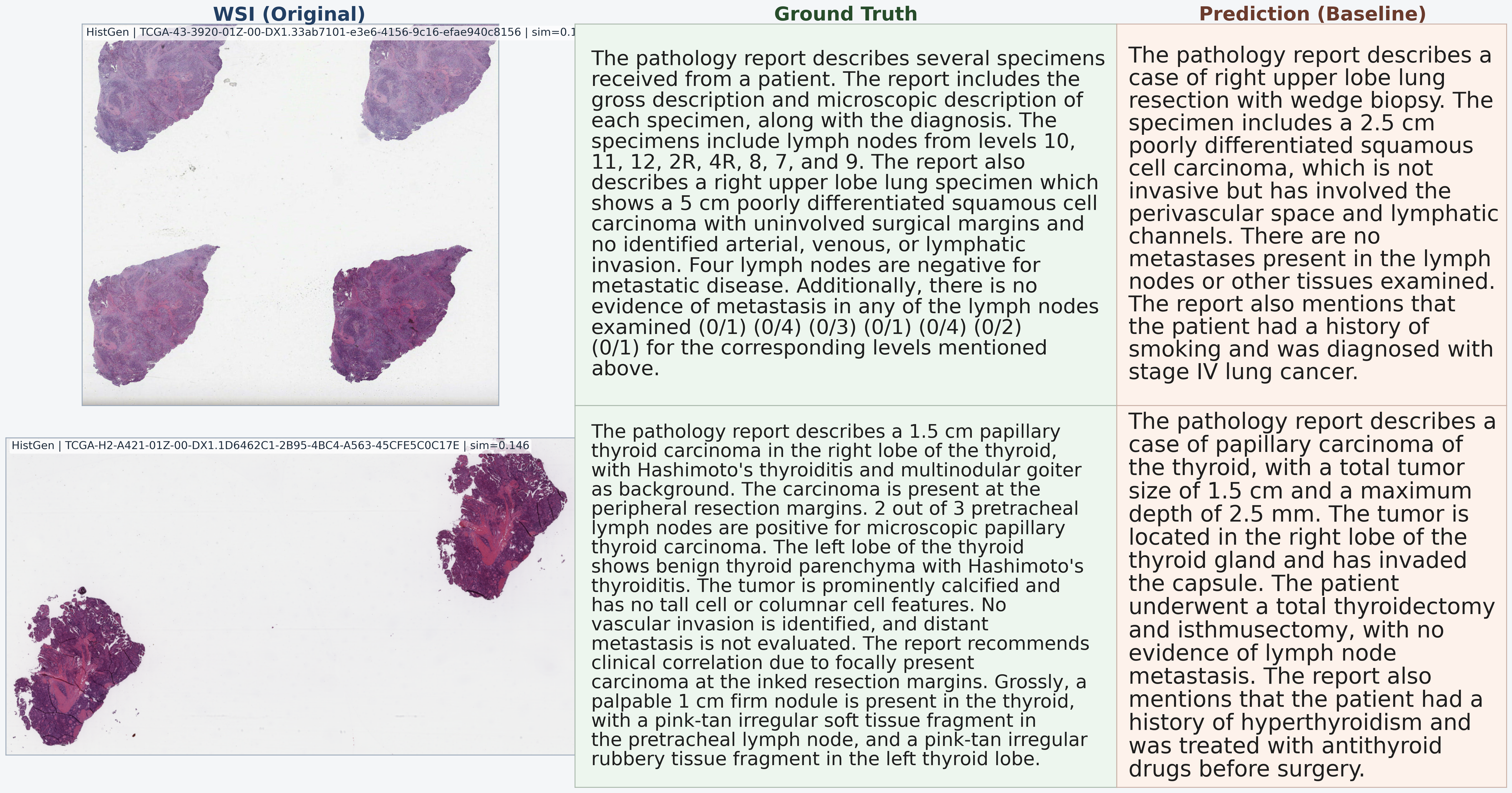}
\caption{\textbf{Generated examples for stage~1 single-WSI captioning task on the HistGen dataset.} From left to right, each row shows the input WSI thumbnail, the reference text, and the Stage~1 baseline prediction.}
\label{fig:stage1_qualitative_histgen}
\end{figure}

\begin{figure}[ht!]
\centering
\includegraphics[width=\linewidth]{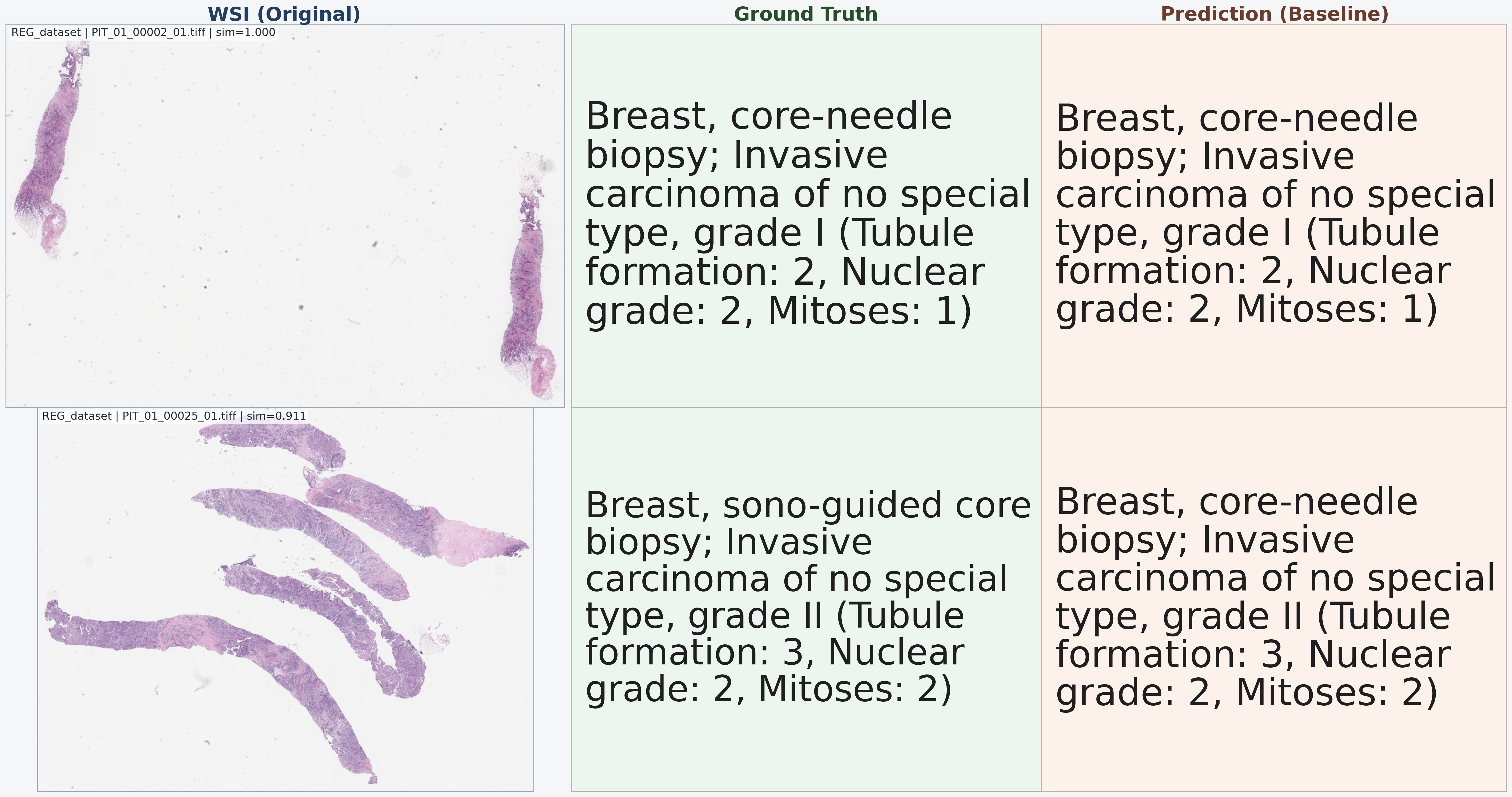}
\caption{\textbf{Generated examples for stage~1 single-WSI captioning task on the REG2025 dataset.} From left to right, each row shows the input WSI thumbnail, the reference text, and the Stage~1 baseline prediction.}
\label{fig:stage1_qualitative_reg}
\end{figure}



Figures~\ref{fig:stage1_qualitative_histgen} and~\ref{fig:stage1_qualitative_reg} show that Stage~1 predictions generally capture specimen type and the main diagnostic statement. HistGen \cite{Guo_HistGen_MICCAI2024} outputs tend to be narrative and descriptive, whereas REG2025 \cite{REG2025} outputs are shorter and more diagnosis-focused. This difference explains why overlap metrics are easier to satisfy on standardized targets and motivates complementing automatic metrics with qualitative and expert review.

\begin{figure}[ht!]
\centering
\includegraphics[width=\linewidth]{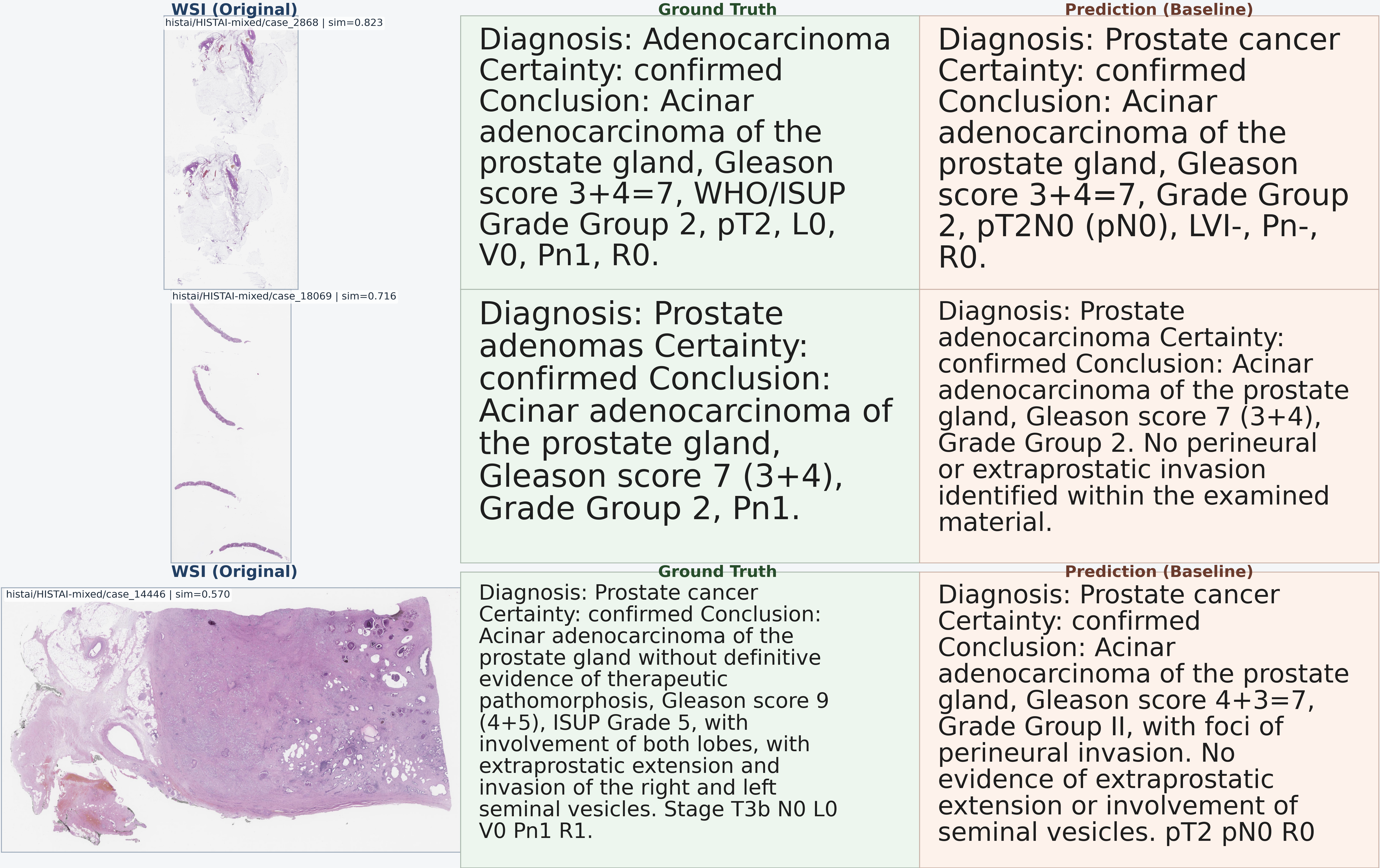}
\caption{\textbf{Generated examples for stage~2 case-level multi-WSI reports on the HistAI dataset.} From left to right, each case shows the packed WSI thumbnails, the reference structured fields, and the Stage~2 baseline prediction in the required schema (\texttt{Diagnosis}, \texttt{Certainty}, \texttt{Conclusion}).}
\label{fig:stage2_qualitative_histai}
\end{figure}

Figure~\ref{fig:stage2_qualitative_histai} shows that the Stage~2 baseline usually follows the required \texttt{Diagnosis}, \texttt{Certainty}, and \texttt{Conclusion} schema even with multiple WSIs. The pathologist audit also noted cases where generated conclusions were clinically interpretable, but some reference conclusions appear to depend on information not fully determined from WSIs alone. This is consistent with the current training setup, which uses visual tokens plus the \texttt{Diagnosis} and \texttt{Conclusion} fields but does not include clinical history, gross findings, or ancillary tests.

\begin{figure}[h]
\centering
\includegraphics[width=\linewidth]{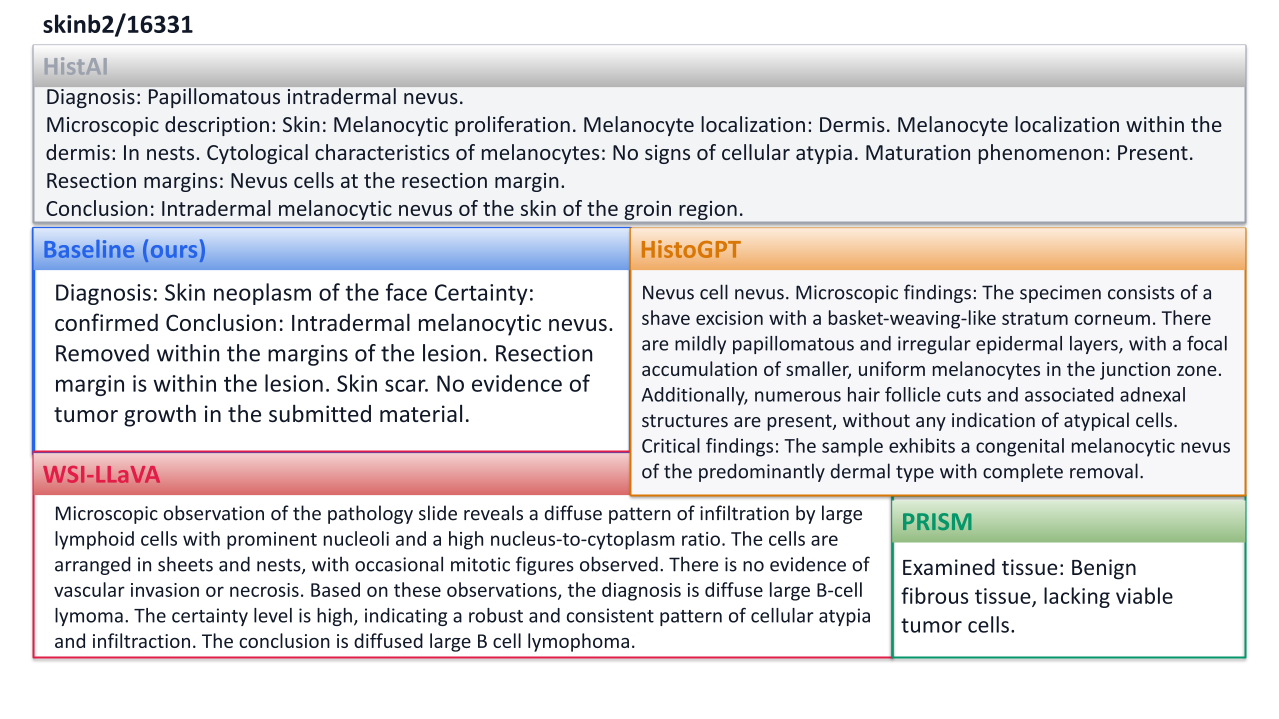}
\caption{\textbf{Example report from our Stage~2 baseline, HistoGPT, WSI-LLaVA, and PRISM.} The reference case corresponds to an intradermal melanocytic nevus. The baseline preserves the main nevus-related conclusion under the required schema. HistoGPT~\cite{Tran2025HistoGPT} provides richer dermatopathology-style prose, whereas WSI-LLaVA \cite{wsi-llava} and PRISM \cite{prism} are less consistent with the reference diagnosis in this example.}
\label{fig:stage2_quantative_histai}
\end{figure}

\textbf{AI-Based Preference Comparison.}
\label{subsec:rubric_eval}
Since the compared models follow different reporting conventions, we conducted an additional comparative evaluation using ChatGPT \cite{openai_chatgpt_2026}. The rubric assessed diagnosis, microscopic description, and final conclusion from the HISTAI \cite{nechaev2025histaiopensourcelargescaleslide} reference fields. We first randomly sampled $600$ cases. After feature extraction and report generation, WSI-LLaVA \cite{wsi-llava} was evaluated on $515$ cases with available \texttt{micro protocol} references, PRISM \cite{prism} on $186$ diagnosis-comparable cases, and HistoGPT \cite{Tran2025HistoGPT} on $100$ skin cases according to its original design. Table~\ref{tab:rubric_eval_summary} summarizes the preference outcomes, and Figure~\ref{fig:stage2_quantative_histai} shows one skin case example from different models.
\begin{table}[h!]
\caption{\textbf{AI-based output preference evaluation.} The table reports the number of cases in which our Stage~2 baseline was preferred, the competing model was preferred, or the two outputs were judged as ties. Each comparison retained the largest valid case set allowed by model output constraints and available reference targets.}
\label{tab:rubric_eval_summary}
\centering
\small
\setlength{\tabcolsep}{4pt}
\begin{tabular}{l c c c c}
\hline
Comparison & Cases & Ours favored & Comparator favored & Tie \\
\hline
Baseline vs WSI-LLaVA & 515 & 460 & 20 & 35 \\
Baseline vs HistoGPT & 100 & 8 & 76 & 16 \\
Baseline vs PRISM & 186 & 176 & 2 & 8 \\
\hline
\end{tabular}
\end{table}

The Stage~2 baseline is strongly favored over WSI-LLaVA \cite{wsi-llava} and PRISM \cite{prism} in the AI-based preference comparison, suggesting that structured Stage~2 supervision improves schema adherence and case-level consistency. The HistoGPT \cite{Tran2025HistoGPT} comparison shows a different pattern. HistoGPT \cite{Tran2025HistoGPT} is preferred in 76 of 100 skin cases, likely because it is specialized for dermatopathology and generates richer microscopic descriptions. The pathologist's audit provides a complementary view. When evaluation is restricted to the common target of conclusion diagnosis, both systems remain imperfect, but our model has a modest advantage in mean diagnostic consistency and pairwise composite wins. In contrast, PRISM \cite{prism} typically produces short diagnostic outputs with limited supporting detail, which explains why the Stage~2 baseline is favored in most PRISM comparisons. Overall, these comparisons provide evidence beyond lexical overlap, but they are not a substitute for expert clinical validation. Reporting style, domain specialization, and reference-field choice all influence the observed preference results.
\section{Ablation Studies}
\label{sec:ablation}
We assess how sensitive the proposed two-stage pipeline is to our design choices. First, we evaluate the design choice that affects both stages, then we perform stage-specific analysis. For interpretability, each ablation modifies one factor relative to the baseline, and we report performance at the checkpoint with the lowest validation loss for each run.

\subsection{Prompt repetition}
\label{subsec:ablation_prompt_both}

\begin{table}[h]
\caption{\textbf{Prompt repetition across both stages.} Stage~1 compares Baseline vs. A1, and Stage~2 compares Baseline vs. B1. BLEU-4 is sacreBLEU normalized to 0--1.}
\label{tab:ablation_prompt_both}
\centering
\small
\setlength{\tabcolsep}{4pt}
\resizebox{\linewidth}{!}{
\begin{tabular}{|l|c|c|c|c|c|c|}
\hline
Setting & ROUGE-L $\uparrow$ & METEOR $\uparrow$ & BLEU-4 $\uparrow$ & BERT F1 $\uparrow$ & Diag (relaxed) $\uparrow$ & Certainty $\uparrow$ \\
\hline
Stage~1 Baseline     & 0.4743 & 0.4810 & 0.1247 & 0.4253 & -- & -- \\
Stage~1 A1 (repeat)  & \textbf{0.4861} & \textbf{0.4997} & \textbf{0.1344} & \textbf{0.4413} & -- & -- \\
\hline
Stage~2 Baseline     & 0.2168 & 0.1772 & 0.0383 & 0.2711 & 0.3000 & \textbf{0.9000} \\
Stage~2 B1 (repeat)  & \textbf{0.2495} & \textbf{0.1988} & \textbf{0.0525} & \textbf{0.3018} & \textbf{0.3333} & \textbf{0.9000} \\
\hline
\end{tabular}
}
\end{table}

Prompt repetition simply duplicates the input prompt and feeds it into the model. Recent work has shown that repeating the instruction can strengthen conditioning in causal language models and improve non-reasoning generation while preserving the target format \cite{leviathan2025promptrepetitionimprovesnonreasoning}. Table \ref{tab:ablation_prompt_both} shows the effect of prompt repetition on generation tasks in both stages. Under the same training and evaluation protocol, prompt repetition consistently improves text metrics based on overlap. In Stage~2, it additionally increases the relaxed diagnosis match, while the certainty score remains unchanged. However, the observed gains remain limited. In our setting, prompts are already short and unambiguous, and Stage~2 supervision explicitly optimizes schema adherence, which limits the benefit of further prompt-based refinement. In addition, a substantial fraction of errors is driven by content uncertainty in the visual evidence or ambiguity in the target text, which repetition cannot resolve. 

\subsection{Stage~1 Ablations}
\label{subsec:ablation_stage1}

\begin{figure}[h]
\centering
\includegraphics[width=\linewidth]{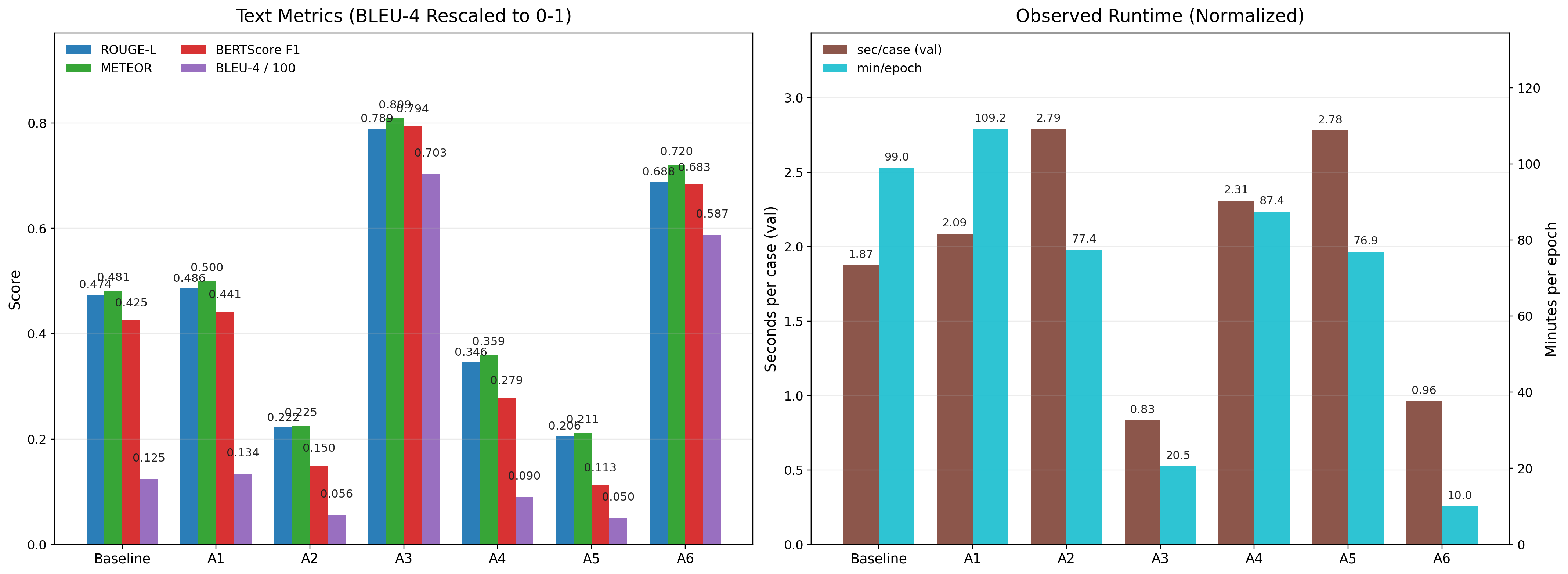}
\caption{\textbf{Stage~1 ablation summary} across text metrics and observed runtime.}
\label{fig:stage1_ablation_summary}
\end{figure}

For Stage~1, we additionally ablate input magnification and training data mixture. The magnification ablation measures the effect of morphological detail under fixed patch size, while the dataset mixture ablation separates the effects of HistGen \cite{Guo_HistGen_MICCAI2024} and REG2025 \cite{REG2025} supervision styles. Table~\ref{tab:ablation_stage1_mag}, Table~\ref{tab:ablation_stage1_mix}, and Figure~\ref{fig:stage1_ablation_summary} summarize the results.

\begin{table*}[ht]
\caption{\textbf{Stage~1 ablation configurations.} $\checkmark$ indicates that the corresponding training dataset is included. \textbf{Baseline} uses a mixture of HistGen and REG2025 dataset at $5\times$ magnification with $512\times 512$ patch size and a single prompt. \textbf{A1} investigates the effect of prompt repetition by repeating the input text prompt two times. \textbf{A2} and \textbf{A3} investigate the effect of dataset choices by training on HistGen-only and REG2025-only, respectively. \textbf{A4} investigates the effect of input patch magnification by extracting patches at $1\times$ magnification. \textbf{A5} and \textbf{A6} further investigate the combination of dataset mixture and low magnification patches on HistGen-only and REG2025-only at $1\times$ magnification.}
\label{tab:ablation_settings_stage1}
\centering
\small
\setlength{\tabcolsep}{4pt}
\resizebox{\linewidth}{!}{
\begin{tabular}{|l|c|c|c|l|r|r|r|r|}
\hline
Setting & Magnification & HistGen & REG2025 & Prompt & \#Train & \#Patches (Train) & \#Val & \#Patches (Val) \\
\hline
Baseline & $5\times$ & $\checkmark$ & $\checkmark$ & single & 11\,859 & 1\,898\,170 & 2\,974 & 476\,023 \\
A1       & $5\times$ & $\checkmark$ & $\checkmark$ & double & 11\,859 & 1\,898\,170 & 2\,974 & 476\,023 \\
A2       & $5\times$ & $\checkmark$ &              & single & 5\,957  & 1\,421\,092 & 1\,499 & 357\,599 \\
A3       & $5\times$ &              & $\checkmark$ & single & 5\,901  & 465\,053   & 1\,476 & 116\,322 \\
A4       & $1\times$ & $\checkmark$ & $\checkmark$ & single & 8\,444  & 79\,990    & 2\,107 & 19\,960 \\
A5       & $1\times$ & $\checkmark$ &              & single & 5\,950  & 75\,525    & 1\,496 & 18\,989 \\
A6       & $1\times$ &              & $\checkmark$ & single & 2\,482  & 15\,238    & 623   & 3\,825 \\
\hline
\end{tabular}
}
\end{table*}

\textbf{(a) Magnification.}
\label{subsec:ablation_stage1_magnification}
We ablate the input magnification to measure the effect of morphological detail, keeping the patch pixel size fixed at $512\times512$. Reducing magnification from $5\times$ to $1\times$ reduces all Stage~1 text metrics, indicating that alignment benefits from preserving more histological detail. Together with Table~\ref{tab:stage1_eff_only}, this supports our use of $5\times$ as a practical middle point. $1\times$ is too coarse for reliable visual-language alignment, while dense $20\times$ patching is substantially more expensive in patch count, preprocessing time, and inference time.

\begin{table}[h]
\caption{\textbf{Stage~1 magnification ablation under single-WSI inputs.} We compare the default $5\times$ setting (Baseline) with the $1\times$ variant (A4) while keeping the patch size fixed at $512\times512$, using the same HistGen+REG supervision mixture and single-instruction prompting. ROUGE-L, METEOR, BLEU-4, and BERTScore-F1 measure text similarity between generated outputs and references.}
\label{tab:ablation_stage1_mag}
\centering
\small
\setlength{\tabcolsep}{4pt}
\begin{tabular}{|l|c|c|c|c|c|}
\hline
Setting & Mag. & ROUGE-L $\uparrow$ & METEOR $\uparrow$ & BLEU-4 $\uparrow$ & BERT F1 $\uparrow$ \\
\hline
Baseline & $5\times$ & \textbf{0.4743} & \textbf{0.4810} & \textbf{0.1247} & \textbf{0.4253} \\
A4 & $1\times$ & 0.3462 & 0.3592 & 0.0903 & 0.2787 \\
\hline
\end{tabular}
\end{table}

\textbf{(b) Training dataset mixture.}
\label{subsec:ablation_stage1_mixture}
We isolate the effect of supervision source by training Stage~1 with three alternatives, HistGen-only (A2), REG2025-only (A3), and their mixture, all at $5\times$ magnification with single-instruction prompting. REG2025-only supervision produces much higher overlap-based metrics than HistGen-only, with the mixed supervision falling in between. This reflects difference in reference style. REG2025 targets are concise diagnostic summaries with recurring, semi-templated phrasing which n-gram overlap metrics reward. In contrast, HistGen references are derived from full report PDFs, then cleaned and summarized, resulting in greater lexical and stylistic variability that overlap-based metrics penalizes even when clinical content is similar. We use mixed supervision as the default to balance standardized diagnostic phrasing with more diverse report-style language, noting that metric gains partly reflect differences in reference structure.

\begin{table}[h]
\caption{\textbf{Stage~1 training dataset mixture ablation at $5\times$ under single-WSI inputs.} We compare training on the mixed supervision set (HistGen+REG2025, Baseline) against single-source training on HistGen only (A2) and REG2025 only (A3), while keeping the patch size fixed at $512\times512$ and using single-instruction prompting. ROUGE-L, METEOR, BLEU-4, and BERTScore-F1 measure text similarity between generated outputs and references.}
\label{tab:ablation_stage1_mix}
\centering
\small
\setlength{\tabcolsep}{4pt}
\resizebox{\linewidth}{!}{
\begin{tabular}{|l|c|c|c|c|}
\hline
Setting & ROUGE-L $\uparrow$ & METEOR $\uparrow$ & BLEU-4 $\uparrow$ & BERT F1 $\uparrow$ \\
\hline
Baseline (mix) & 0.4743 & 0.4810 & 0.1247 & 0.4253 \\
A2 (HistGen) & 0.2220 & 0.2245 & 0.0564 & 0.1498 \\
A3 (REG2025) & \textbf{0.7893} & \textbf{0.8087} & \textbf{0.7035} & \textbf{0.7936} \\
\hline
\end{tabular}
}
\end{table}

\subsection{Stage~2 Ablations}
\label{subsec:ablation_stage2}

\begin{figure}[h]
\centering
\includegraphics[width=\linewidth]{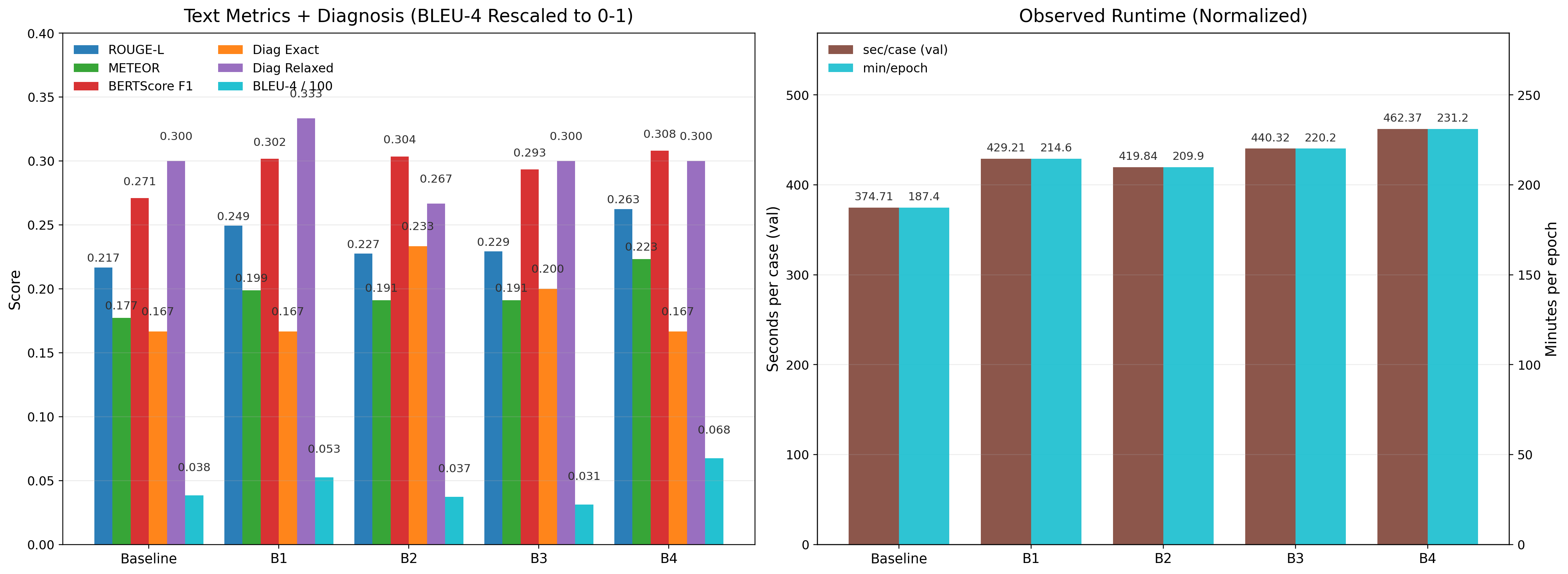}
\caption{\textbf{Stage~2 ablation summary} for case-level structured reporting with multi-WSI packing. The figure compares text metrics, field-level correctness, and average runtime across the baseline and B1--B4 settings. For each setting, results correspond to the checkpoint with the lowest validation loss.}
\label{fig:stage2_ablation_perf_runtime}
\end{figure}

For Stage~2, we evaluate explicit WSI boundary markers and vision-token dropout. Boundary markers test whether slide segmentation cues improve packed multi-WSI generation, while dropout tests whether token-level regularization improves robustness. Table~\ref{tab:ablation_stage2_marker} and Table~\ref{tab:ablation_stage2_dropout} summarize the performance impact of each factor, and Figure~\ref{fig:stage2_ablation_perf_runtime} compares combined metrics and runtime across Stage~2 settings.

\begin{table}[ht]
\caption{\textbf{Stage~2 ablation settings.} Each setting modifies a single factor relative to the Stage~2 baseline. The \textbf{baseline} uses HistAI \cite{nechaev2025histaiopensourcelargescaleslide} at $5\times$ magnification with a patch size of $512\times512$, a single prompt, no WSI marker token, and no vision token dropout. \textbf{B1:} Investigates the effect of prompt repetition in case-level report generation. \textbf{B2:} Investigates the effect of vision-token dropout regularization in long packed visual sequences. \textbf{B3} and \textbf{B4:} Investigate the effect of index-specific WSI marker tokens and a single WSI marker token as explicit slide-boundary cues during generation.}
\label{tab:ablation_settings_stage2}
\centering
\scriptsize
\setlength{\tabcolsep}{5pt}
\begin{tabular}{|l|l|l|}
\hline
Setting & Category & Change vs. baseline \\
\hline
Baseline & baseline & -- \\
B1 & prompt repetition & duplicate the system instruction \\
B2 & vision-token dropout & \texttt{vision\_token\_dropout}: 0.0 $\rightarrow$ 0.2 \\
B3 & WSI marker token & use index-specific WSI marker tokens for each WSI \\
B4 & WSI marker token & use the same WSI marker token for all WSIs \\
\hline
\end{tabular}
\end{table}

\textbf{(a) WSI marker token.}
\label{subsec:ablation_stage2_marker}
In Stage~2, where multiple WSIs are combined into a single sequence, explicit boundary cues may help maintain slide attribution during structured generation. We test the following two settings. First, we use a set of WSI marker tokens unique to each WSI index within a case (B3). Second, we use a single WSI marker token to separate all WSIs within a case (B4). Table~\ref{tab:ablation_stage2_marker} reports the comparison. B3 gives small improvements in ROUGE-L, METEOR, and BERTScore, but does not improve BLEU-4. B4 achieves the highest overall text metrics and improves certainty match. This suggests that index-specific WSI marker tokens can introduce noise, as marker tokens for early indices are trained more often than those for later indices, making the latter less effective. Using a single WSI marker embedding makes the boundary marker the main segmentation cue, which sharpens slide attribution and improves both text metrics and certainty match.

\begin{table}[ht]
\caption{\textbf{Stage~2 WSI marker ablation under case-level multi-WSI inputs.} We evaluate whether explicit slide-boundary cues improve structured report generation by comparing the baseline without WSI marker tokens against two marker variants. B3 inserts a unique marker token before each WSI segment, and B4 inserts the same marker token for all WSI segments. ROUGE-L, METEOR, BLEU-4, and BERTScore-F1 measure text similarity between generated reports and references, while Diag (relaxed) and Certainty measure field-level correctness under the structured schema.}
\label{tab:ablation_stage2_marker}
\centering
\small
\setlength{\tabcolsep}{4pt}
\resizebox{\linewidth}{!}{
\begin{tabular}{|l|c|c|c|c|c|c|}
\hline
Setting & ROUGE-L $\uparrow$ & METEOR $\uparrow$ & BLEU-4 $\uparrow$ & BERT F1 $\uparrow$ & Diag (relaxed) $\uparrow$ & Certainty $\uparrow$ \\
\hline
Baseline (No WSI marker token) & 0.2168 & 0.1772 & 0.0383 & 0.2711 & \textbf{0.3000} & 0.9000 \\
B3 (Unique token per WSI) & 0.2292 & 0.1912 & 0.0312 & 0.2934 & \textbf{0.3000} & 0.9000 \\
B4 (Same WSI marker token for all WSIs) & \textbf{0.2625} & \textbf{0.2232} & \textbf{0.0676} & \textbf{0.3081} & \textbf{0.3000} & \textbf{0.9333} \\
\hline
\end{tabular}
}
\end{table}

\textbf{(b) Vision-token dropout.}

We evaluate vision-token dropout in B2 as a robustness intervention for long packed visual sequences in Stage~2. During training, a fixed proportion of vision tokens is randomly dropped, serving as token-level regularization. Compared to the baseline, token dropout increases BERTScore-F1 but does not improve BLEU-4 and reduces the relaxed diagnosis match. This suggests that the current dropout setting favours sentence-level semantic similarity but removes visual evidence that is important for diagnosis-related fields. Overall, vision-token dropout does not provide a consistent benefit in this configuration and likely requires more careful tuning of rate and schedule to avoid loss of clinically relevant signals.

\label{subsec:ablation_stage2_dropout}
\begin{table}[ht]
\caption{\textbf{Stage~2 vision-token dropout ablation under case-level multi-WSI inputs.} We evaluate token-level regularization for long packed visual sequences by comparing the baseline with B2, where 20\% of vision tokens are randomly dropped during training. }
\label{tab:ablation_stage2_dropout}
\centering
\small
\setlength{\tabcolsep}{4pt}
\resizebox{\linewidth}{!}{
\begin{tabular}{|l|c|c|c|c|c|c|}
\hline
Setting & ROUGE-L $\uparrow$ & METEOR $\uparrow$ & BLEU-4 $\uparrow$ & BERT F1 $\uparrow$ & Diag (relaxed) $\uparrow$ & Certainty $\uparrow$ \\
\hline
Baseline & 0.2168 & 0.1772 & \textbf{0.0383} & 0.2711 & \textbf{0.3000} & \textbf{0.9000} \\
B2 (dropout 0.2) & \textbf{0.2275} & \textbf{0.1910} & 0.0373 & \textbf{0.3036} & 0.2667 & \textbf{0.9000} \\
\hline
\end{tabular}
}
\end{table}

\section{Limitations and Conclusion}
We presented a simple token-efficient VLM for case-level pathology synoptic report generation under constrained computational resources. The model combines a frozen pathology patch encoder, a lightweight MLP aligner, explicit WSI boundary tokens, and a Qwen2.5-3B-Instruct decoder trained with a two-stage supervised recipe. Our results show that this deliberately minimal architecture can serve as a practical baseline for multi-WSI report generation when paired with low-token visual inputs and memory-efficient training.

Several limitations remain. First, 5× patching provides a favourable efficiency-detail trade-off, but it may miss fine cytologic details that require high-magnification review. Second, the two-layer MLP aligner is intentionally simple and may not match the reasoning capacity of cross-attention modules, Perceiver-style bottlenecks, or dedicated slide/case encoders. Third, report generation from WSIs alone is inherently limited because some reference conclusions may depend on clinical metadata, gross findings, immunohistochemistry, or other non-image evidence. Finally, overlap-based metrics and AI-based judging are imperfect proxies for clinical correctness. Although the pathologist audit provides a more clinically grounded assessment, it is limited to 67 overlapping skin cases, a conclusion-level target, and a small-scale expert review; a larger, blinded multi-pathologist evaluation is needed before drawing clinical conclusions.

Overall, this work establishes a reproducible resource-aware baseline for case-level pathology report generation. Future work should investigate multi-scale token selection, stronger evidence grounding, clinically targeted evaluation, and broader pathologist review.

\subsubsection{\ackname} The authors used ChatGPT for English writing consistency and fluency during the preparation of this manuscript. The content has been reviewed and edited by the authors as needed after using the AI tool. The authors take full responsibility for the content of the published article.

\subsubsection{\discintname} The authors declare no competing interests.

\bibliographystyle{splncs04}
\bibliography{references}

@InProceedings{Guo_HistGen_MICCAI2024,
        author = { Guo, Zhengrui and Ma, Jiabo and Xu, Yingxue and Wang, Yihui and Wang, Liansheng and Chen, Hao},
        title = { { HistGen: Histopathology Report Generation via Local-Global Feature Encoding and Cross-modal Context Interaction } },
        booktitle = {proceedings of Medical Image Computing and Computer Assisted Intervention -- MICCAI 2024},
        year = {2024},
        publisher = {Springer Nature Switzerland},
        volume = {LNCS 15004},
        month = {October},
        page = {189 -- 199}
}

@misc{REG2025,
  author       = {{Grand Challenge}},
  title        = {Report Generation in Pathology using Pan-Asia Giga-pixel WSIs (REG2025)},
  year         = {2025},
  howpublished = {\url{https://reg2025.grand-challenge.org/}},
  note         = {-MICCAI Registered Challenge. DOI: 10.5281/zenodo.15081613 (accessed 2026-02-24).}
}

@misc{nechaev2025histaiopensourcelargescaleslide,
      title={HISTAI: An Open-Source, Large-Scale Whole Slide Image Dataset for Computational Pathology}, 
      author={Dmitry Nechaev and Alexey Pchelnikov and Ekaterina Ivanova},
      year={2025},
      eprint={2505.12120},
      archivePrefix={arXiv},
      primaryClass={eess.IV},
      url={https://arxiv.org/abs/2505.12120}, 
}

@article{Tran2025HistoGPT,
  title   = {Generating Dermatopathology Reports from Gigapixel Whole Slide Images with HistoGPT},
  author  = {Tran, Manuel and
             Schmidle, Paul and
             Guo, Ruifeng Ray and
             Wagner, Sophia J. and
             Koch, Valentin and
             Lupperger, Valerio and
             Novotny, Brenna and
             Murphree, Dennis H. and
             Hardway, Heather D. and
             D’Amato, Marina and
             Lefkes, Judith and
             Geijs, Daan J. and
             Feuchtinger, Annette and
             Böhner, Alexander and
             Kaczmarczyk, Robert and
             Biedermann, Tilo and
             Amir, Avital L. and
             Mooyaart, Antien L. and
             Ciompi, Francesco and
             Litjens, Geert and
             Wang, Chen and
             Comfere, Nneka I. and
             Eyerich, Kilian and
             Braun, Stephan A. and
             Marr, Carsten and
             Peng, Tingying},
  journal = {Nature Communications},
  volume  = {16},
  number  = {1},
  pages   = {4886},
  year    = {2025},
  doi     = {10.1038/s41467-025-60014-x},
  issn    = {2041-1723}
}

@inproceedings{papineni-etal-2002-bleu,
    title = "{B}leu: a Method for Automatic Evaluation of Machine Translation",
    author = "Papineni, Kishore  and
      Roukos, Salim  and
      Ward, Todd  and
      Zhu, Wei-Jing",
    editor = "Isabelle, Pierre  and
      Charniak, Eugene  and
      Lin, Dekang",
    booktitle = "Proceedings of the 40th Annual Meeting of the Association for Computational Linguistics",
    month = jul,
    year = "2002",
    address = "Philadelphia, Pennsylvania, USA",
    publisher = "Association for Computational Linguistics",
    doi = "10.3115/1073083.1073135",
    pages = "311--318"
}

@inproceedings{lin-2004-rouge,
    title = "{ROUGE}: A Package for Automatic Evaluation of Summaries",
    author = "Lin, Chin-Yew",
    booktitle = "Text Summarization Branches Out",
    month = jul,
    year = "2004",
    address = "Barcelona, Spain",
    publisher = "Association for Computational Linguistics",
    url = "https://aclanthology.org/W04-1013/",
    pages = "74--81"
}

@inproceedings{lavie-agarwal-2007-meteor,
    title = "{METEOR}: An Automatic Metric for {MT} Evaluation with High Levels of Correlation with Human Judgments",
    author = "Lavie, Alon  and
      Agarwal, Abhaya",
    editor = "Callison-Burch, Chris  and
      Koehn, Philipp  and
      Fordyce, Cameron Shaw  and
      Monz, Christof",
    booktitle = "Proceedings of the Second Workshop on Statistical Machine Translation",
    month = jun,
    year = "2007",
    address = "Prague, Czech Republic",
    publisher = "Association for Computational Linguistics",
    url = "https://aclanthology.org/W07-0734/",
    pages = "228--231"
}

@inproceedings{Zhang2020BERTScore,
title={BERTScore: Evaluating Text Generation with BERT},
author={Tianyi Zhang and Varsha Kishore and Felix Wu and Kilian Q. Weinberger and Yoav Artzi},
booktitle={International Conference on Learning Representations},
year={2020},
url={https://openreview.net/forum?id=SkeHuCVFDr}
}

@article{alagha2026atlaspatch,
  title   = {AtlasPatch: Efficient Tissue Detection and High-throughput Patch Extraction for Computational Pathology at Scale},
  author  = {Alagha, Ahmed and Leclerc, Christopher and Kotp, Yousef and Metwally, Omar and Moras, Calvin and Rentopoulos, Peter and Rostami, Ghodsiyeh and Nguyen, Bich Ngoc and Baig, Jumanah and Khellaf, Abdelhakim and Trinh, Vincent Quoc-Huy and Mizouni, Rabeb and Otrok, Hadi and Bentahar, Jamal and Hosseini, Mahdi S.},
  journal = {arXiv preprint arXiv:2602.03998},
  year    = {2026}
}

@article{lu2024conch,
  title={A visual-language foundation model for computational pathology},
  author={Lu, Ming Y and Chen, Bowen and Williamson, Drew FK and Chen, Richard J and Liang, Ivy and Ding, Tong and Jaume, Guillaume and Odintsov, Igor and Le, Long Phi and Gerber, Georg and others},
  journal={Nature Medicine},
  pages={863–-874},
  volume={30},
  year={2024},
  publisher={Nature Publishing Group}
}

@misc{qwen2025qwen25technicalreport,
      title={Qwen2.5 Technical Report}, 
      author={Qwen and : and An Yang and Baosong Yang and Beichen Zhang and Binyuan Hui and Bo Zheng and Bowen Yu and Chengyuan Li and Dayiheng Liu and Fei Huang and Haoran Wei and Huan Lin and Jian Yang and Jianhong Tu and Jianwei Zhang and Jianxin Yang and Jiaxi Yang and Jingren Zhou and Junyang Lin and Kai Dang and Keming Lu and Keqin Bao and Kexin Yang and Le Yu and Mei Li and Mingfeng Xue and Pei Zhang and Qin Zhu and Rui Men and Runji Lin and Tianhao Li and Tianyi Tang and Tingyu Xia and Xingzhang Ren and Xuancheng Ren and Yang Fan and Yang Su and Yichang Zhang and Yu Wan and Yuqiong Liu and Zeyu Cui and Zhenru Zhang and Zihan Qiu},
      year={2025},
      eprint={2412.15115},
      archivePrefix={arXiv},
      primaryClass={cs.CL},
      url={https://arxiv.org/abs/2412.15115}, 
}

@misc{leviathan2025promptrepetitionimprovesnonreasoning,
      title={Prompt Repetition Improves Non-Reasoning LLMs}, 
      author={Yaniv Leviathan and Matan Kalman and Yossi Matias},
      year={2025},
      eprint={2512.14982},
      archivePrefix={arXiv},
      primaryClass={cs.LG},
      url={https://arxiv.org/abs/2512.14982}, 
}

@inproceedings{
lora,
title={Lo{RA}: Low-Rank Adaptation of Large Language Models},
author={Edward J Hu and Yelong Shen and Phillip Wallis and Zeyuan Allen-Zhu and Yuanzhi Li and Shean Wang and Lu Wang and Weizhu Chen},
booktitle={International Conference on Learning Representations},
year={2022},
url={https://openreview.net/forum?id=nZeVKeeFYf9}
}

@misc{llava,
      title={Visual Instruction Tuning}, 
      author={Liu, Haotian and Li, Chunyuan and Wu, Qingyang and Lee, Yong Jae},
      publisher={NeurIPS},
      year={2023},
}

@article{Gelu,
  title={Gaussian error linear units (gelus)},
  author={Hendrycks, Dan and Gimpel, Kevin},
  journal={arXiv preprint arXiv:1606.08415},
  year={2016}
}

@article{titan,
  title={A multimodal whole-slide foundation model for pathology},
  author={Ding, Tong and Wagner, Sophia J and Song, Andrew H and Chen, Richard J and Lu, Ming Y and Zhang, Andrew and Vaidya, Anurag J and Jaume, Guillaume and Shaban, Muhammad and Kim, Ahrong and others},
  journal={Nature Medicine},
  pages={1--13},
  year={2025},
  publisher={Nature Publishing Group US New York}
}

@article{gigapath,
  title={A whole-slide foundation model for digital pathology from real-world data},
  author={Xu, Hanwen and Usuyama, Naoto and Bagga, Jaspreet and Zhang, Sheng and Rao, Rajesh and Naumann, Tristan and Wong, Cliff and Gero, Zelalem and González, Javier and Gu, Yu and Xu, Yanbo and Wei, Mu and Wang, Wenhui and Ma, Shuming and Wei, Furu and Yang, Jianwei and Li, Chunyuan and Gao, Jianfeng and Rosemon, Jaylen and Bower, Tucker and Lee, Soohee and Weerasinghe, Roshanthi and Wright, Bill J. and Robicsek, Ari and Piening, Brian and Bifulco, Carlo and Wang, Sheng and Poon, Hoifung},
  journal={Nature},
  year={2024},
  publisher={Nature Publishing Group UK London}
}

@article{cpath,
title = {Computational pathology: A survey review and the way forward},
journal = {Journal of Pathology Informatics},
volume = {15},
pages = {100357},
year = {2024},
issn = {2153-3539},
doi = {https://doi.org/10.1016/j.jpi.2023.100357},
url = {https://www.sciencedirect.com/science/article/pii/S2153353923001712},
author = {Mahdi S. Hosseini and Babak Ehteshami Bejnordi and Vincent Quoc-Huy Trinh and Lyndon Chan and Danial Hasan and Xingwen Li and Stephen Yang and Taehyo Kim and Haochen Zhang and Theodore Wu and Kajanan Chinniah and Sina Maghsoudlou and Ryan Zhang and Jiadai Zhu and Samir Khaki and Andrei Buin and Fatemeh Chaji and Ala Salehi and Bich Ngoc Nguyen and Dimitris Samaras and Konstantinos N. Plataniotis},
}

@InProceedings{wsi-llava,
    author    = {Liang, Yuci and Lyu, Xinheng and Chen, Wenting and Ding, Meidan and Zhang, Jipeng and He, Xiangjian and Wu, Song and Xing, Xiaohan and Yang, Sen and Wang, Xiyue and Shen, Linlin},
    title     = {WSI-LLaVA: A Multimodal Large Language Model for Whole Slide Image},
    booktitle = {Proceedings of the IEEE/CVF International Conference on Computer Vision (ICCV)},
    month     = {October},
    year      = {2025},
    pages     = {22718-22727}
}

@article{prism,
  title={PRISM: A Multi-Modal Generative Foundation Model for Slide-Level Histopathology},
  author={Shaikovski, George and Casson, Adam and Severson, Kristen and Zimmermann, Eric and Wang, Yi Kan and Kunz, Jeremy D and Retamero, Juan A and Oakley, Gerard and Klimstra, David and Kanan, Christopher and others},
  journal={arXiv preprint arXiv:2405.10254},
  year={2024}
}

@misc{longnet,
      title={LongNet: Scaling Transformers to 1,000,000,000 Tokens}, 
      author={Jiayu Ding and Shuming Ma and Li Dong and Xingxing Zhang and Shaohan Huang and Wenhui Wang and Nanning Zheng and Furu Wei},
      year={2023},
      eprint={2307.02486},
      archivePrefix={arXiv},
      primaryClass={cs.CL},
      url={https://arxiv.org/abs/2307.02486}, 
}

@inproceedings{blip2,
      title={{BLIP-2:} Bootstrapping Language-Image Pre-training with Frozen Image Encoders and Large Language Models}, 
      author={Junnan Li and Dongxu Li and Silvio Savarese and Steven Hoi},
      year={2023},
      booktitle={ICML},
}

@article{train-short,
  title={Train short, test long: Attention with linear biases enables input length extrapolation},
  author={Press, Ofir and Smith, Noah A and Lewis, Mike},
  journal={arXiv preprint arXiv:2108.12409},
  year={2021}
}

@article{Tomczak2015TheCG,
  title={The Cancer Genome Atlas (TCGA): an immeasurable source of knowledge},
  author={Katarzyna Tomczak and Patrycja Czerwińska and Maciej Wiznerowicz},
  journal={Contemporary Oncology},
  year={2015},
  volume={19},
  pages={A68 - A77},
  url={https://api.semanticscholar.org/CorpusID:12829250}
}

@InProceedings{pathttt,
author="He, Haoyu
and Hosseini, Mahdi S.
and Wang, Yang",
editor="Oguz, Ipek
and Zhang, Shaoting
and Metaxas, Dimitris N.",
title="PathTTT: Test-Time Training with Meta-auxiliary Learning for Pathology Image Classification",
booktitle="Information Processing in Medical Imaging",
year="2026",
publisher="Springer Nature Switzerland",
address="Cham",
pages="33--46",
isbn="978-3-031-96628-6"
}

@InProceedings{prompt,
author="Sharma, Vasudev
and Alagha, Ahmed
and Khellaf, Abdelhakim
and Trinh, Vincent Quoc-Huy
and Hosseini, Mahdi S.",
editor="Hadjali, Allel
and Maiorana, Emanuele
and Gusikhin, Oleg
and Sansone, Carlo",
title="Investigating Zero-Shot Diagnostic Pathology in Vision-Language Models with Efficient Prompt Design",
booktitle="Deep Learning Theory and Applications",
year="2025",
publisher="Springer Nature Switzerland",
address="Cham",
pages="263--279",
isbn="978-3-032-04339-9"
}

@misc{openai_chatgpt_2026,
  author       = {{OpenAI}},
  title        = {ChatGPT},
  year         = {2026},
  howpublished = {\url{https://openai.com/chatgpt/}},
  note         = {AI chatbot}
}
%




\end{document}